\documentclass[a4paper,fleqn]{cas-sc}

\usepackage[authoryear]{natbib}

\usepackage{booktabs}   % For professional tables
\usepackage{amssymb}    % Provides \checkmark
\usepackage{pifont}     % Provides \ding{51} (check) and \ding{55} (cross)
\usepackage{array}
\usepackage{subcaption}

\makeatletter
\@fleqntrue
\makeatother

\usepackage{etoolbox}
\AtBeginEnvironment{keywords}{%
    \vspace{-10pt} % Reduce space before
    \setlength{\parskip}{2pt} % Reduce space between items
}
\AtEndEnvironment{keywords}{%
    \vspace{-5pt} % Reduce space after
}

\usepackage{caption}
\usepackage{graphicx}

\usepackage[switch]{lineno}
\usepackage{hyperref}

\hypersetup{
    colorlinks=true,
    linkcolor=[rgb]{0,0.60,0.75},
    citecolor=[rgb]{0,0.60,0.75},
    urlcolor=[rgb]{0,0.60,0.75}
}

\def\tsc#1{\csdef{#1}{\textsc{\lowercase{#1}}\xspace}}
\tsc{WGM}
\tsc{QE}
\tsc{EP}
\tsc{PMS}
\tsc{BEC}
\tsc{DE}
\begin{document}
\sloppy
\let\WriteBookmarks\relax
\def\floatpagepagefraction{1}
\def\textpagefraction{.001}
\shorttitle{HTC-SGA Former}
\shortauthors{Rayan Merghani Ahmed et~al.}

\title [mode = title]{HTC-SGA Former: A Hybrid Transformer-CNN Network with Self-Guided Attention and a New Boundary-Weighted Adaptive Loss for Coronary DSA Vessel Segmentation}

\affiliation[1]{organization={Shenzhen Institutes of Advanced Technology (SIAT)},
                addressline={Chinese Academy of Sciences (CAS)}, 
                city={Shenzhen},
                postcode={518055}, 
                country={China}}
\affiliation[2]{organization={University of Chinese Academy of Sciences (UCAS)},
                city={Beijing},
                postcode={101408}, 
                country={China}}
\affiliation[3]{organization={Department of Biomedical Engineering and Systems, Faculty of Engineering},
                addressline={Cairo University},
                city={Cairo},
                postcode={12613}, 
                country={Egypt}}

\author[1,2]{Rayan Merghani Ahmed}
\ead{rayan@siat.ac.cn}
\author[1,2]{Marwa Omer Mohammed Omer}
\ead{marwaomer@siat.ac.cn}

\author[3]{Mohamed Elmanna}
\ead{Mohamed_elmanna@hotmail.com}

\author[1]{Shijie Li}
\ead{sj.li5@siat.ac.cn}
\author[1]{Bin Li}
\ead{b.li2@siat.ac.cn}

\author[1]{Shoujun Zhou}[orcid= 0000-0003-3232-6796]
\cormark[1]
\ead{sj.zhou@siat.ac.cn}

\cortext[cor1]{Corresponding author}

\begin{abstract}
Accurate coronary Digital Subtraction Angiography (DSA) vessel segmentation is essential for computer-aided diagnosis and treatment planning of coronary artery disease (CAD). However, thin low-contrast vessels, background interference, and severe vessel-background class imbalance make reliable segmentation of weak distal branches and vessel boundaries challenging. Existing segmentation methods struggle to balance global contextual reasoning with preservation of weak distal vessels, vessel continuity, and fine boundaries. To address these limitations, we propose HTC-SGA Former, a lightweight hybrid Transformer-CNN framework for coronary DSA vessel segmentation. The framework employs a CNN encoder for local vessel morphology extraction and a Transformer decoder for contextual feature modeling. A Multi-Scale Global-Local Window Attention (MS-GLWA) block performs efficient global-local contextual modeling, while a Self-Guided Feature Attention (SGFA) module enhances weak-vessel responses. In addition, a Boundary-Weighted Adaptive Compound Loss (BWACL) emphasizes thin-vessel boundaries and adaptively balances vessel recovery and boundary refinement under severe class imbalance. Experimental results on private right and left coronary artery DSA subsets demonstrate that HTC-SGA Former consistently outperforms 14 state-of-the-art segmentation methods while maintaining a compact architecture with only 0.81M parameters. Moreover, BWACL improves performance over conventional binary cross-entropy and Dice losses across four widely used encoder-decoder architectures, demonstrating strong cross-backbone applicability. HTC-SGA Former provides accurate and parameter-efficient coronary DSA vessel segmentation by improving thin-vessel recovery, vessel continuity, and boundary localization through complementary global-local contextual modeling, vessel-focused refinement, and adaptive optimization. The proposed framework enables reliable and computationally efficient coronary vessel analysis, supporting quantitative image analysis and future computer-assisted cardiovascular interventions.  
\end{abstract}


\maketitle

\section{Introduction}
Coronary artery disease (CAD) remains one of the leading causes of morbidity and mortality worldwide, with prevalence estimates around 10\% in the general population \citep{xia2020early,frkak2022pathophysiology}. CAD is characterized by atherosclerotic changes in the coronary arteries that impair myocardial blood supply \citep{frkak2022pathophysiology,pagliaro2020myocardial}. Early detection is essential for timely intervention and prevention of acute myocardial infarction, where advanced imaging modalities play a critical role in identifying subclinical disease \citep{xia2020early}. Among these modalities, Digital Subtraction Angiography (DSA), particularly X-ray coronary angiography (XCA), is widely used for visualization of coronary vessels and stenotic lesions due to its real-time imaging capability and detailed vascular assessment \citep{frkak2022pathophysiology,zeng2019automatic}. Furthermore, advances in image processing and vessel segmentation have enhanced the clinical utility of DSA through automated analysis \citep{frkak2022pathophysiology,moccia2018blood}. However, accurate coronary vessel segmentation remains challenging because of low contrast, motion artifacts, complex vascular structures, and severe class imbalance \citep{deng2025multi,algarni2022multi}. While manual delineation is labor-intensive and subjective, traditional image processing techniques often fail under these challenging conditions. Consequently, deep learning (DL) has emerged as the dominant paradigm for coronary vessel segmentation due to its superior accuracy and efficiency in semantic segmentation tasks \citep{nobre2023coronary}.
\begin{figure}
	\centering
	\includegraphics[width=0.8\textwidth]{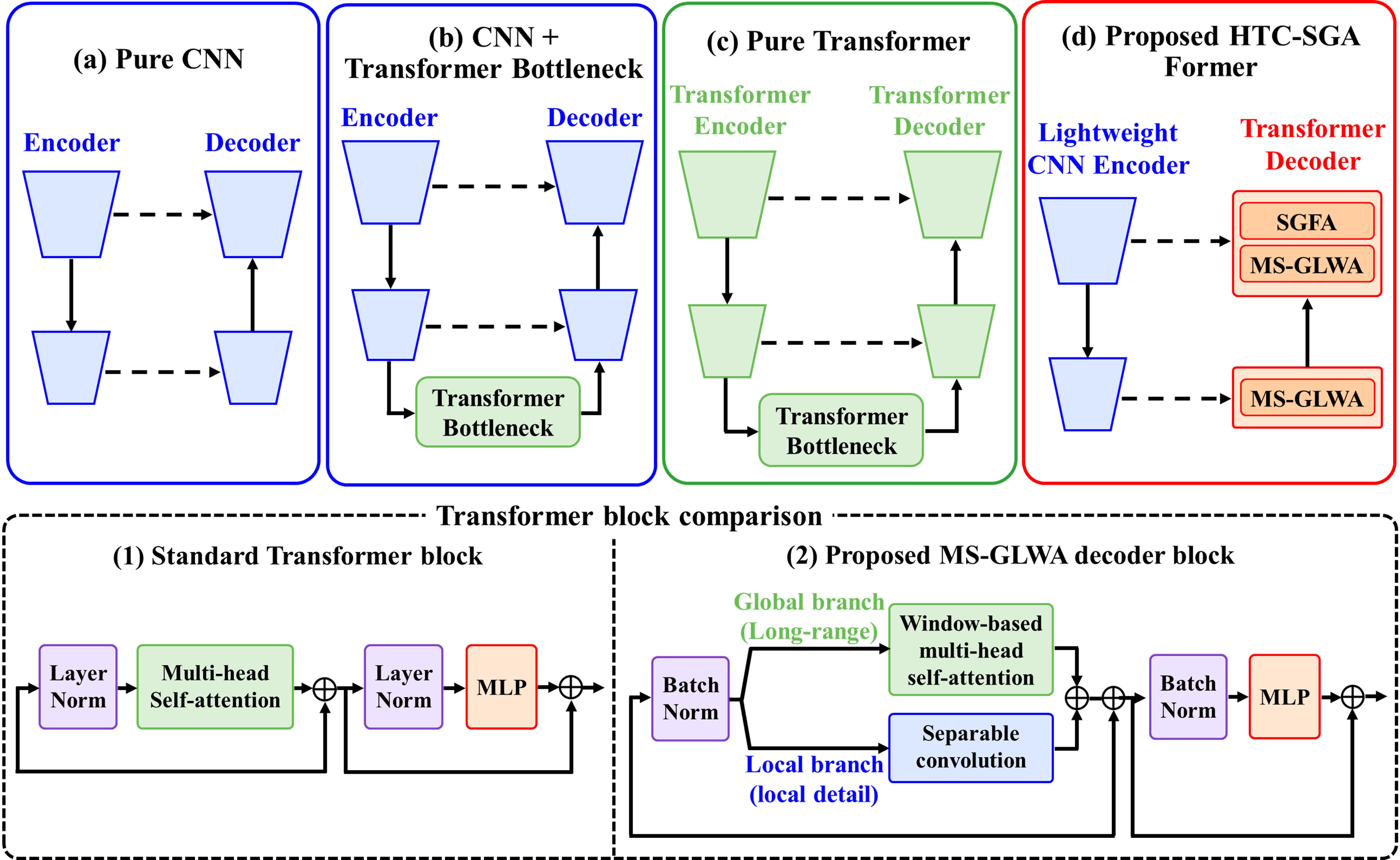}
	\caption{Comparison of medical image segmentation architectures: (a) pure CNN encoder-decoder (U-Net \citep{ronneberger2015u}); (b) hybrid CNN with Transformer bottleneck (e.g., TransUNet \citep{chen2021transunet}); (c) pure Transformer encoder-decoder (e.g., SwinUNet \citep{cao2022swin}); and (d) the HTC-SGA Former combining a lightweight CNN encoder with a Transformer decoder. The lower panel compares the standard Transformer block with the proposed MS-GLWA decoder block integrating global window-based self-attention and local  convolution.}
	\label{fig1}
	\vspace{-11pt} 
\end{figure}
\\
Coronary vessel segmentation in XCA and DSA has evolved from handcrafted enhancement methods to deep encoder–decoder, topology-aware, and self-/semi-supervised learning frameworks. Early approaches relied on foreground extraction, clustering, and vessel enhancement techniques. Fu et al. \citep{fu2024robust} employed TV-TRPCA with region growing, while Mardani et al. \citep{mardani2025segmentation} combined DBSCAN with morphological reconstruction. Hybrid pipelines were later explored, where Algarni et al. \citep{algarni2022multi} proposed a multi-stage CAD framework incorporating preprocessing, attention-based segmentation, and classification. Despite improved robustness, these methods depend on handcrafted components, limiting end-to-end learning. With the advancement of deep learning, most studies shifted toward supervised encoder–decoder architectures based on U-Net and FCN. Yang et al. \citep{yang2019deep} demonstrated major-vessel segmentation using deep U-Net backbones, while Zhu et al. \citep{zhu2021coronary} improved multi-scale contextual modeling with PSPNet. Subsequent works introduced lightweight and attention-enhanced architectures, including VSSC Net \citep{samuel2021vssc}, and lightweight attention-based model by Chang et al. \citep{chang2024optimizing}. However, many of these methods process frames independently and remain less effective for thin vessels. To improve segmentation under low-contrast conditions, several studies incorporated attention mechanisms, multi-scale fusion, and contrast-aware modeling. CIDN \citep{zhang2024cidn} introduced bio-inspired attention and edge-aware optimization, while DFA-Net, and MDA-Net \citep{deng2024dfa,deng2025multi} leveraged multi-scale spatial–channel features for improved small-vessel delineation. Similarly, MSA-UNet3+ with supervised prototypical contrastive loss (SPCL) was proposed to enhance discriminative feature learning and alleviate class imbalance in coronary DSA segmentation \citep{ahmed2026msa}. Although these methods enhance boundary quality, they remain largely appearance-driven and still struggle with extremely fine distal vessels under low-SNR conditions. To further preserve vascular continuity, later studies introduced geometric supervision and progressive refinement strategies. Zhang et al. \citep{zhang2023centerline} proposed centerline-supervised U-Net with channel attention, while Progressive Perception Learning (PPL) \citep{zhang2022progressive} and ITG-Net \citep{gao2026iterative} improved continuity through contextual refinement and temporal guidance. Nevertheless, these methods often increase model complexity and may remain biased toward dominant vessel structures. Another research direction aims to reduce dependence on dense annotations. Pu et al. \citep{pu2023semi} proposed a semi-supervised framework based on cross-teaching and consistency learning using an Inception-SwinUnet backbone, while Challier et al. \citep{challier2025cm} further introduced CM-UNet with contrastive learning and masked reconstruction. Despite reduced annotation requirements, accurately modeling fine vessel topology under low-SNR conditions and complex coronary variability remains challenging.

However, accurately modeling global anatomical context while preserving thin-vessel continuity remains challenging. CNN-dominant methods mainly capture local patterns due to restricted receptive fields, limiting their ability to distinguish vessels from overlapping anatomy and background structures. Although attention mechanisms partially alleviate this limitation, dense global self-attention incurs high computational cost. Moreover, severe vessel-background class imbalance, low contrast, and anatomical overlap make thin distal vessels and boundary regions difficult to preserve, often resulting in fragmented vessel continuity. To address these limitations, we propose HTC-SGA Former, a hybrid Transformer-CNN framework that combines a lightweight CNN encoder for local vessel representation with a Transformer decoder for global-local contextual reasoning. As illustrated in the upper part of Fig.~\ref{fig1}, the proposed framework differs from pure CNN, bottleneck-based hybrid, and pure Transformer architectures by placing the Transformer within the decoder to model long-range anatomical dependencies for vessel continuity. Furthermore, the lower part of Fig.~\ref{fig1} compares the standard Transformer block with the proposed Multi-Scale Global-Local Window Attention (MS-GLWA) decoder block, highlighting the integration of global and local feature modeling in our design. In addition, a Self-Guided Feature Attention (SGFA) module is introduced to refine weak vessel responses under low-contrast conditions. Moreover, a boundary-weighted adaptive compound loss (BWACL) is introduced to address severe class imbalance and improve thin-vessel boundary preservation. The main contributions are summarized as follows:

\begin{enumerate}
    \item HTC-SGA Former, a lightweight hybrid CNN--Transformer framework, is proposed for coronary DSA vessel segmentation to address limited global contextual modeling, weak-vessel representation, and severe vessel-background class imbalance. The framework combines a CNN encoder and a Transformer decoder for efficient global-local feature learning.
    \item A Multi-Scale Global-Local Window Attention (MS-GLWA) block is introduced to improve vessel continuity recovery and weak-vessel representation. By integrating multi-scale window-based self-attention with separable convolution, MS-GLWA captures long-range dependencies while preserving fine vessel structures and connectivity.
    \item A Self-Guided Feature Attention (SGFA) module is designed to perform vessel-focused feature refinement. Through preliminary vessel guidance and residual attention, SGFA enhances weak vessel responses and suppresses background interference, improving discrimination of low-contrast and distal vessel structures.
    \item A Boundary-Weighted Adaptive Compound Loss (BWACL) is proposed to address severe vessel-background class imbalance. By integrating boundary-aware weighting with adaptive balancing of weighted focal and Dice losses, BWACL improves thin-vessel recovery and boundary localization.
    \item Extensive experiments on private right and left coronary artery DSA subsets validate the effectiveness of HTC-SGA Former. The proposed framework consistently outperforms 14 state-of-the-art segmentation methods while maintaining high parameter efficiency. Furthermore, BWACL demonstrates strong cross-backbone applicability by improving performance over conventional binary cross-entropy and Dice losses across four widely used encoder--decoder architectures.
\end{enumerate}

\section{Related work}
\label{sec:related}
Medical image segmentation has evolved from traditional algorithms to deep learning frameworks. However, accurate segmentation of challenging structures such as coronary DSA vessels requires preservation of fine vessel structures and global anatomical continuity. To address these challenges, recent studies have mainly focused on architectural innovation and training optimization. CNN-based methods effectively capture local features but are limited in modeling long-range dependencies, whereas Transformer-based approaches improve global contextual reasoning at the cost of higher computational complexity and reduced spatial-detail preservation. Consequently, hybrid CNN–Transformer frameworks have emerged to balance local feature preservation and global contextual modeling. In parallel, loss-function optimization has become increasingly important for improving boundary delineation and structural continuity.

\subsection{Traditional CNN-Based Methods}
Convolutional Neural Networks (CNNs) have become the dominant paradigm in medical image segmentation following the introduction of the U-Net encoder–decoder architecture with skip connections by Ronneberger et al. \citep{ronneberger2015u}. The success of U-Net led to numerous variants, including U-Net++ \citep{Zongwei_Rahman_Nima_Jianming_2022}, Attention U-Net \citep{oktay2018attention}, Res2UNet \citep{he2016deep}, and TransUNet \citep{chen2021transunet}, which improved feature representation, and attention modeling.
In coronary X-ray angiography segmentation, CNN-based encoder–decoder models replaced handcrafted vessel enhancement with end-to-end feature learning, progressively improving robustness and automation. Shen et al. \citep{shen2023dbcu} proposed DBCU-Net for improved coronary segmentation accuracy, while Park et al. \citep{park2023selective} introduced selective ensemble learning for major-vessel ICA segmentation. Molenaar et al. \citep{molenaar2025deep} further explored automated segmentation pipelines, and multicenter learning strategies to improve clinical applicability and generalization. In parallel, Pan et al. \citep{pan2021coronary} demonstrated the importance of imbalance-aware training for vessel recovery. Despite these advances, CNN-based methods still struggle with global vessel-tree modeling, weak distal branches, and severe vessel–background class imbalance, motivating the development of attention- and Transformer-based architectures.

\subsection{Attention Mechanisms and Transformer-Based Methods}
Attention mechanisms were introduced to enhance CNNs by enabling models to focus on relevant features while suppressing irrelevant information. Woo et al. \citep{woo2018cbam} proposed the Convolutional Block Attention Module (CBAM), which applies channel and spatial attention sequentially, improving feature representation in segmentation tasks. In medical imaging, Oktay et al. \citep{oktay2018attention} extended this concept through attention gates in U-Net.
Building on attention mechanisms, Transformer architectures have gained prominence for modeling long-range dependencies through self-attention, which is important for elongated, discontinuous, or globally distributed anatomy. In medical image segmentation literature, Liu et al. \citep{liu2023mestrans} proposed MESTrans, which uses spatial Transformers to strengthen global-context modeling while reducing the encoder–decoder semantic gap, whereas Wu et al. \citep{wu2024hd} developed HD-Former to capture hierarchical dependencies and long-range feature interaction. To improve multi-scale Transformer segmentation, Huang et al. \citep{huang2024hst} proposed HST-MRF, which uses heterogeneous Swin Transformer, and Du et al. \citep{li2023cpftransformer} designed CPFTransformer to fuse Transformer reasoning with a context pyramid for stronger region-level discrimination. In coronary angiography, Huang et al. \citep{huang2025deep} adapted MedSAM and VM-UNet for vessel segmentation, but their results also suggest that thin vessels and stenotic boundaries still require task-specific refinement mechanisms. Despite their strong global modeling capability, these transformer-based approaches introduce high computational cost and data dependency, and often lack fine-grained boundary precision due to limited local feature modeling. Additionally, pure Transformer models may struggle with small vessel detection and edge continuity, especially in low-contrast angiographic images. To address these issues, recent works have explored hybrid architectures that combine CNNs and Transformers, leveraging the strengths of both approaches.

\subsection{Hybrid CNN–Transformer Method and Loss Function Design}
Recent research has increasingly favored hybrid CNN–Transformer architectures and specialized loss functions, reflecting the view that architecture alone is insufficient for clinically reliable segmentation. Fu et al. \citep{fu2024hmsu} proposed HmsU-Net, combining CNN locality with Transformer multi-scale reasoning to address the incomplete feature learning of single-paradigm models, while Li et al. \citep{li2023transu2} introduced TransU\(^2\)-Net, and Tang et al. \citep{tang2023combined} combined a deformable model with a medical Transformer, both arguing that hybridization can reduce the spatial-detail loss seen in Transformer-heavy designs while overcoming the limited receptive field of standard CNNs. In coronary angiography, Wang et al. \citep{wang2024coronary} used UT-BTNet as a hybrid local–global architecture and coupled it with boundary aggregation and topology-preserving supervision. This shift toward supervision-aware modeling is also consistent with general medical segmentation studies: Wang et al. \citep{wang2022boundary} developed a boundary-aware context neural network, and Qiu et al. \citep{qiu2022dynamic} developed a dynamic boundary-insensitive loss to stabilize training near ambiguous interfaces. More recently, Peng et al. \citep{peng2024boundary} emphasized boundary-aware information maximization in self-supervised segmentation, and Wang et al. \citep{wang2025dpgnet} proposed DPGNet, a boundary-aware framework driven by uncertainty perception, confirming that clinically relevant segmentation quality often depends on explicitly steering the optimizer toward boundaries, structure, and hard pixels. Collectively, these studies indicate that the current state of the field is not simply a transition from CNNs to Transformers, but rather a move toward jointly designed local-global architectures and objective functions that can better handle class imbalance, structural continuity, and fine-vessel delineation in difficult medical images, including coronary angiography.

Overcoming these limitations is essential for clinically reliable coronary DSA vessel segmentation. In particular, accurate delineation of thin distal vessels requires a framework that can jointly capture broader anatomical context, preserve fine local vessel details, and optimize learning under severe vessel-background imbalance. To this end, this study proposes HTC-SGA Former, a hybrid Transformer-CNN network that combines efficient global-local context modeling with vessel-focused feature refinement. In addition, a boundary-weighted adaptive compound loss is introduced to emphasize thin-vessel boundaries, mitigate class imbalance, and improve the balance between vessel detection and boundary localization. Through this joint architectural and optimization design, the proposed method is aims to improve thin-vessel continuity and boundary accuracy in challenging coronary DSA images.

\begin{figure}
	\centering
	\includegraphics[width=0.80\textwidth]{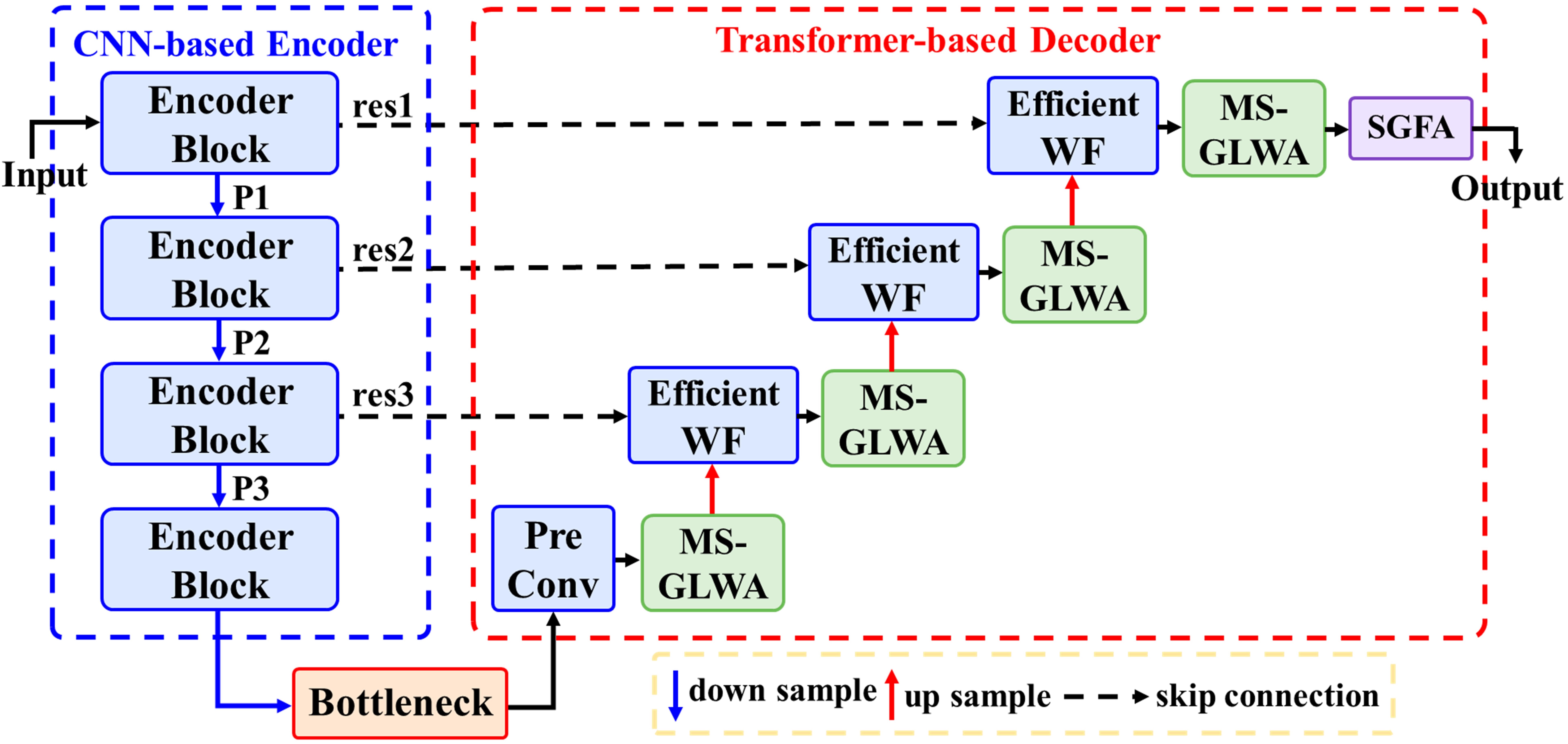}
	\caption{Overall architecture of HTC-SGA Former, comprising a lightweight CNN encoder and a Transformer-based decoder incorporating Efficient WF, MS-GLWA, and SGFA for coronary vessel segmentation.}
	\label{fig2}
	\vspace{-11pt} 
\end{figure}

\section{Method}
\label{sec:method}
\subsection{Overview}
In this study, we propose HTC-SGA Former for coronary DSA vessel segmentation to address three key challenges: limited global anatomical context, weak representation of thin distal vessels, and severe vessel-background class imbalance. The proposed framework combines a lightweight CNN encoder with a Transformer decoder, as illustrated in Fig.~\ref{fig2}, to preserve fine local vessel details in the encoding stage and progressively enhance global-local contextual reasoning during decoding. In particular, the MS-GLWA block enables efficient global-local context modeling, while the SGFA module refines weak and low-contrast vessel features through self-guided attention and multi-scale context aggregation. Furthermore, a boundary-weighted adaptive compound loss (BWACL) is introduced to emphasize thin-vessel boundaries and improve optimization under class imbalance. Together, these components are designed to enhance vessel continuity, thin-vessel delineation, and boundary accuracy in challenging coronary DSA images.

\subsection{CNN-based encoder}
Within HTC-SGA Former, the CNN encoder serves as a lightweight front-end feature extractor that preserves fine vessel morphology while progressively generating hierarchical features for decoding. The encoder consists of four convolutional stages, each followed by spatial down-sampling. Given an input coronary DSA image $X \in \mathbb{R}^{B \times 1 \times H \times W}$, feature extraction is performed using convolutional blocks denoted by $E_i(\cdot)$, where each block comprises two consecutive $3 \times 3$ convolutional layers with batch normalization and ReLU activation, followed by dropout (0.1). A $2 \times 2$ max-pooling operation with stride $2$ is applied after each block for down-sampling. The hierarchical feature extraction process is formulated in Eq.~(\ref{eq1})--Eq.~(\ref{eq4}):
\vspace{-3pt}
\begin{align}
\mathrm{F}_1 &= \mathrm{E}_1(\mathrm{X}) \in \mathbb{R}^{\mathrm{B} \times 48 \times \mathrm{H} \times \mathrm{W}} \label{eq1} \\
\mathrm{F}_2 &= \mathrm{E}_2(\operatorname{Pool}(\mathrm{F}_1)) \in \mathbb{R}^{\mathrm{B} \times 48 \times \mathrm{H} / 2 \times \mathrm{W} / 2} \label{eq2} \\
\mathrm{F}_3 &= \mathrm{E}_3(\operatorname{Pool}(\mathrm{F}_2)) \in \mathbb{R}^{\mathrm{B} \times 48 \times \mathrm{H} / 4 \times \mathrm{W} / 4} \label{eq3} \\
\mathrm{F}_4 &= \mathrm{E}_4(\operatorname{Pool}(\mathrm{F}_3)) \in \mathbb{R}^{\mathrm{B} \times 48 \times \mathrm{H} / 8 \times \mathrm{W} / 8} \label{eq4}
\end{align}

where $E_i$ denotes the $i$-th encoding block, $\operatorname{Pool}(\cdot)$ denotes max pooling, and $B$, $H$, and $W$ represent the batch size and spatial dimensions, respectively. The channel dimension is fixed at $48$ across all encoding stages to maintain computational efficiency while preserving vessel feature representation.

Following the encoding stages, a bottleneck block is applied to the deepest feature map to further enrich high-level semantic representations before decoding, as formulated in Eq.~(\ref{eq5}):
\vspace{-3pt}
\begin{equation}\label{eq5}
\mathrm{F}_{\mathrm{b}}=\mathrm{B}\left(\operatorname{Pool}\left(\mathrm{F}_4\right)\right) \in \mathbb{R}^{\mathrm{B} \times 128 \times \mathrm{H}/16 \times \mathrm{W}/16}
\end{equation}
where $B(\cdot)$ denotes the bottleneck block, consisting of a $3 \times 3$ convolution that expands the channel dimension from $48$ to $128$, followed by batch normalization, ReLU activation, and dropout (0.1). The resulting feature map $F_b$ captures high-level semantic information. The multi-scale feature maps ${F_1,F_2,F_3,F_b}$ with channel dimensions $[48,48,48,128]$ are forwarded to the decoder through skip connections for contextual modeling and vessel refinement.

\begin{figure}
	\centering
	\includegraphics[width=0.75\textwidth]{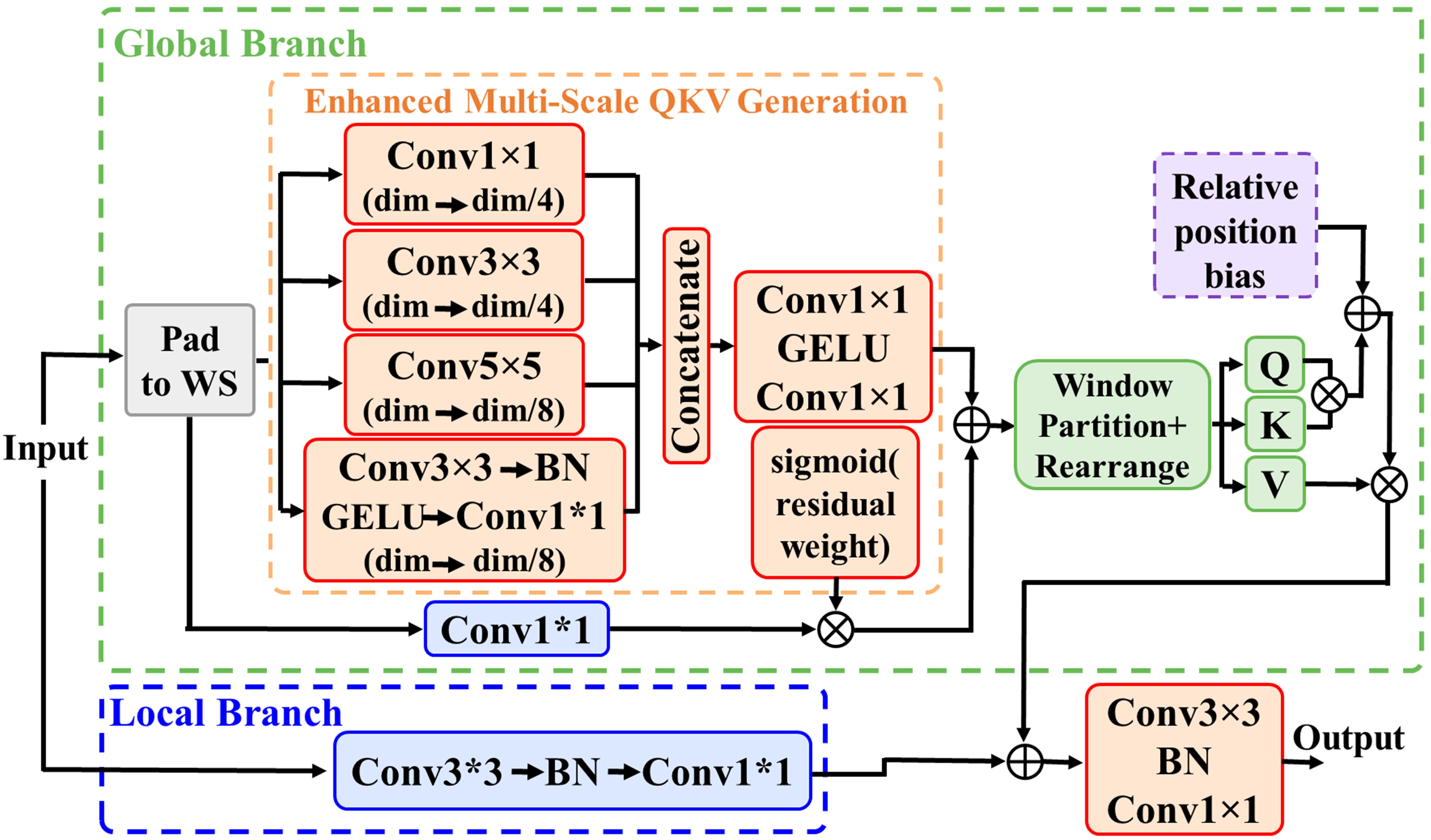}
	\caption{Architecture of the MS-GLWA module. The global branch performs enhanced multi-scale QKV generation and window-based self-attention, while the local branch preserves fine spatial details through convolutions.}
	\label{fig4}
	\vspace{-11pt} 
\end{figure}
\subsection{Transformer-based decoder}
Although attention-guided and hybrid CNN–Transformer methods have improved coronary vessel segmentation \citep{oktay2018attention,woo2018cbam,fu2024hmsu}, balancing broader anatomical context modeling with weak distal-vessel recovery and background-interference suppression in DSA images remains challenging. To address these limitations, HTC-SGA Former employs a Transformer-based decoder that refines hierarchical encoder features through global-local contextual modeling and vessel-focused enhancement. The decoder integrates the proposed Multi-Scale Global-Local Window Attention (MS-GLWA) block and Self-Guided Feature Attention (SGFA) module to improve vessel continuity, distal-branch recovery, and boundary-aware refinement with high computational efficiency.

\subsubsection{Multi-scale global-local window attention block (MS-GLWA)}
Accurate coronary DSA vessel segmentation requires both long-range contextual modeling and preservation of fine vessel structures. While global context supports vessel continuity and improves discrimination between vessels and surrounding structures, local information remains essential for preserving thin-vessel details and sharp boundaries. To address this trade-off, the proposed MS-GLWA block combines global and local branches. As illustrated in Fig.~\ref{fig1}, MS-GLWA consists of a multi-scale enhanced attention module, a multilayer perceptron (MLP), normalization layers, and residual connections. The attention module employs two parallel branches for complementary feature modeling (Fig.~\ref{fig4}). The local branch preserves fine vessel morphology and boundary details using a separable convolution comprising a depthwise $3 \times 3$ convolution, batch normalization, and a pointwise $1 \times 1$ convolution. The resulting local feature map is defined in Eq.~(\ref{eq6}):
\vspace{-3pt}
\begin{equation}\label{eq6} 
\mathrm{X}_{\text {local }}=\operatorname{Conv}_{1 \times 1}\left(\mathrm{BN}\left(\text {Conv}_{\text{depthwise } 3 \times 3}(\mathrm{X})\right)\right) \in \mathbb{R}^{\mathrm{B} \times \mathrm{C} \times \mathrm{H} \times \mathrm{W}}
\end{equation}
In parallel, the global branch captures broader anatomical context through window-based multi-head self-attention. Unlike the standard Swin Transformer \citep{liu2021swin}, which employs a single linear projection, the proposed design generates multi-scale feature representations using four parallel convolutional branches with different receptive fields. These features are then concatenated along the channel dimension as in Eq.~(\ref{eq7}):
\vspace{-2pt} 
\begin{equation}\label{eq7}
\begin{aligned}
& F_{1 \times 1} = \operatorname{Conv}_{1 \times 1}^{C \rightarrow C/4}(X), \quad
F_{3 \times 3} = \operatorname{Conv}_{3 \times 3}^{C \rightarrow C/4}(X), \quad
F_{5 \times 5} = \operatorname{Conv}_{5 \times 5}^{C \rightarrow C/8}(X), \\
& F_{\mathrm{dw}} = \operatorname{Conv}_{1 \times 1}^{C \rightarrow C/8}\left(\operatorname{GELU}\left(\operatorname{BN}\left(\operatorname{Conv}_{\text{depthwise } 3 \times 3}(X)\right)\right)\right), \\
& \mathrm{F}_{\mathrm{cat}}=\left[\mathrm{F}_{1 \times 1}, \mathrm{~F}_{3 \times 3}, \mathrm{~F}_{5 \times 5}, \mathrm{~F}_{\mathrm{dw}}\right] \in \mathbb{R}^{\mathrm{B} \times(3 \mathrm{C} / 4) \times \mathrm{H} \times \mathrm{W}}.
\end{aligned}
\end{equation}
By constructing query $Q$, key $K$, and value $V$ from multi-scale representations, the global branch better captures coronary DSA vascular patterns. The concatenated features are then processed by a lightweight fusion module comprising two $1 \times 1$ convolutional layers with GELU activation, followed by a final $1 \times 1$ convolution to generate the QKV tensor, as defined in Eq.~(\ref{eq9}):
\vspace{-5pt}
\begin{equation}\label{eq9}
\mathrm{F}_{\text {fusion }}=\text { Fusion }\left(\mathrm{F}_{\text {cat }}\right) \in \mathbb{R}^{\mathrm{B} \times(3 \times \mathrm{dqkv}) \times \mathrm{H} \times \mathrm{W}}
\end{equation}
where, $d_{\text{qkv}}=C / r$, and $r$ denotes the reduction ratio which is set to $2$ by default. To preserve the original feature information and stabilize optimization, a residual projection is introduced in Eq.~(\ref{eq10}):
\vspace{-5pt}
\begin{equation}\label{eq10}
\begin{aligned}
& F_{\text{res}} = \operatorname{Conv}_{1 \times 1}^{C \rightarrow 3 d_{qkv}}(X), \\
& QKV_{\text{total}} = F_{\text{fusion}} + \alpha \cdot F_{\text{res}} \in \mathbb{R}^{B \times 3 d_{qkv} \times H \times W}, & \alpha = \sigma\left(\alpha_{\text{param}}\right), \quad \alpha_{\text{param}} \in \mathbb{R}^1.
\end{aligned}
\end{equation}
Here, the learnable residual weight $\alpha$ (initialized to $0.3$ for the right subset and $0.5$ for the left subset) is constrained to $(0,1)$ using the sigmoid function $\sigma$ to ensure stable feature fusion. The fused QKV tensor is then partitioned into non-overlapping windows of size $w_s \times w_s$ and split into $Q$, $K$, and $V$ tensors, as defined in Eq.~(\ref{eq11}):
\vspace{-2pt}
\begin{equation}\label{eq11}
Q, K, V=\operatorname{Split}\left(\text { WindowPartition }\left(Q K V_{\text {total }}\right)\right)
\end{equation}

where each tensor has shape $\mathbb{R}^{(B \times N_w) \times h \times w_s^2 \times d}$, with $N_w=(H/w_s)(W/w_s)$ denoting the number of windows, $h$ the number of heads, and $d=d_{\mathrm{qkv}}/h$ the head dimension. In our implementation, $w_s=8$, $d=16$, and $h=8$. Further details on window-based multi-head self-attention are provided in Swin Transformer \citep{liu2021swin}. However, the fixed scaling factor used in standard self-attention may lead to suboptimal attention distributions especially in low-contrast and noisy coronary DSA images, affecting thin-vessel continuity. To address this limitation, a learnable temperature parameter $\tau$ is introduced to adaptively scale the attention logits before softmax normalization, as expressed in Eq.~(\ref{eq13}):
\vspace{-5pt}
\begin{equation}\label{eq13}
X_{\text {global }}=\operatorname{Rearrange}\left(\operatorname{Softmax}\left(\tau \cdot \frac{Q K^T}{\sqrt{d}}+B\right) V\right) \in \mathbb{R}^{B \times C \times H \times W}
\end{equation}
% \vspace{-5pt}
where, $Q$, $K$, and $V \in \mathbb{R}^{(B\times N_w)\times h \times w_s^2 \times d}$ denote the query, key, and value tensors within each window, and $B$ is the relative position bias derived from a learnable table $\hat{B}\in\mathbb{R}^{(2w_s-1)\times(2w_s-1)\times h}$. The learnable temperature parameter $\tau$, initialized to 0.5, adaptively scales the attention logits to improve robustness in low-contrast and noisy coronary DSA images. Smaller $\tau$ values encourage broader contextual integration, whereas larger values produce sharper vessel-background discrimination. The Rearrange operation restores the windowed representation to the spatial domain, producing the global feature map $X_{\text{global}}$ while preserving the original resolution. Finally, the global and local features are fused through element-wise summation, as in Eq.~(\ref{eq14}):
\vspace{-5pt} 
\begin{equation}\label{eq14}
X_{\text {fused }}=X_{\text {global }}+X_{\text {local }}
\end{equation}
% \vspace{-3pt}
Although MS-GLWA improves global-local reasoning and vessel continuity, accurate coronary DSA segmentation still requires refinement of weak and low-contrast vessel responses. This motivates the SGFA module.

\begin{figure}
	\centering
	\includegraphics[width=0.75\textwidth]{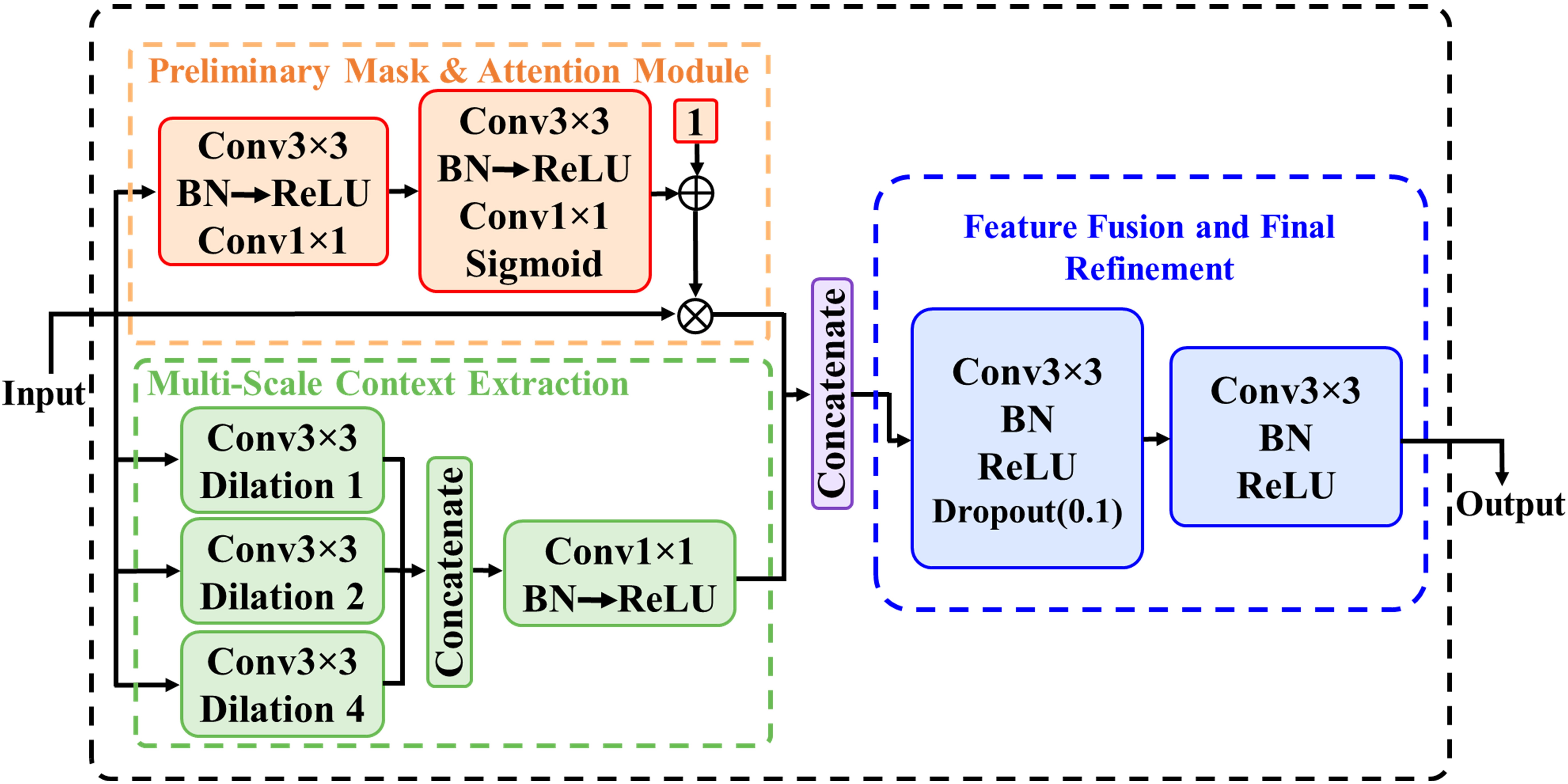}
	\caption{Structure of the SGFA module. The module generates a preliminary vessel-guided attention map, extracts multi-scale contextual features through parallel dilated convolutions, and fuses them for final vessel-focused refinement.}
	\label{fig5}
	\vspace{-11pt} 
\end{figure}
\subsubsection{Self-guided feature attention module (SGFA)}
Although MS-GLWA improves global-local contextual modeling, accurate coronary DSA segmentation still requires refinement of weak and low-contrast vessel responses, particularly thin distal branches that are easily confused with background structures. To address this, the proposed SGFA module performs vessel-focused decoder refinement by generating guidance directly from decoder features. As shown in Fig.~\ref{fig5}, SGFA consists of three components: (1) preliminary vessel-mask generation for self-guided attention, (2) multi-scale context extraction, and (3) feature fusion and refinement for suppressing background interference and enhancing vessel representations.
\subsection*{Preliminary mask generation and attention modulation}
Given the input decoder feature map $F$, SGFA first generates a coarse vessel probability map to estimate vessel regions. This preliminary mask serves as a structural prior for subsequent feature refinement by guiding the network toward candidate vessel locations. The preliminary vessel mask is computed as follows:
\vspace{-3pt}
\begin{equation}\label{eq15}
\mathbf{M}_{p r e}=\operatorname{Conv}_{1 \times 1}\left(\delta\left(\operatorname{BN}\left(\operatorname{Conv}_{3 \times 3}^{C \rightarrow C / 2}(\mathbf{F})\right)\right)\right) \in \mathbb{R}^{B \times 1 \times H \times W}
\end{equation}
where, $\delta$ denotes the ReLU activation and BN represents batch normalization. The resulting mask provides coarse vascular localization for the self-guided attention mechanism. To generate feature-enhancement weights, the mask is expanded to $C/4$ channels and projected to $C$ channels, as expressed in Eq.~(\ref{eq16}):
\vspace{-3pt}
\begin{equation}\label{eq16}
\mathbf{A}=\sigma\left(\operatorname{Conv}_{1 \times 1}^{C / 4 \rightarrow C}\left(\delta\left(\operatorname{BN}\left(\operatorname{Conv}_{3 \times 3}^{1 \rightarrow C / 4}\left(\mathbf{M}_{p r e}\right)\right)\right)\right)\right) \in \mathbb{R}^{B \times C \times H \times W}
\end{equation}
where, $A$ denotes the attention map with values in $[0,1]$, indicating the decoder features to be enhanced at each spatial location. The attention is applied using a residual formulation rather than direct gating, as given in Eq.~(\ref{eq17}):
\vspace{-5pt}
\begin{equation}\label{eq17}
\mathbf{F}_{\text {attended }}=\mathbf{F} \odot(\mathbf{1}+\mathbf{A})
\end{equation}
where, $\odot$ denotes element-wise multiplication. The residual formulation enhances weak vessel responses while preserving the original features. Compared with hard gating, residual attention provides more stable refinement by preserving baseline decoder features while enhancing vessel-consistent responses, improving low-contrast vessel enhancement and weak-signal recovery.

\subsection*{Multi-scale context extraction and fusion}
In addition to self-guided attention, coronary vessel segmentation requires multi-scale refinement due to variations in vessel thickness. To address this, SGFA employs a multi-scale context extraction module based on parallel dilated convolutions. Given the input decoder feature map $F$, three parallel branches capture vascular features at different scales. The resulting features are concatenated along the channel dimension and fused, as formulated in Eq.~(\ref{eq18}):
\vspace{-3pt}
\begin{equation}\label{eq18}
\begin{aligned}
& C_1=\operatorname{Conv}_{3 \times 3}^{C \rightarrow C / 2}(F, d=1),\quad 
C_2=\operatorname{Conv}_{3 \times 3}^{C \rightarrow C / 2}(F, d=2),\quad
C_3=\operatorname{Conv}_{3 \times 3}^{C \rightarrow C / 2}(F, d=4), \\
& \mathbf{C}_{\text {cat }}=\left[\mathbf{C}_1, \mathbf{C}_2, \mathbf{C}_3\right] \in \mathbb{R}^{B \times(3 C / 2) \times H \times W}, \\
& \mathbf{C}_{\text {fused }}=\delta\left(\mathrm{BN}\left(\operatorname{Conv}_{1 \times 1}^{3 C / 2 \rightarrow C}\left(\mathbf{C}_{\text {cat }}\right)\right)\right) \in \mathbb{R}^{B \times C \times H \times W}.
\end{aligned}
\end{equation}
where, $d$ denotes the dilation rate. Dilation rates $1$, $2$, and $4$ capture fine, intermediate, and broader vascular features, respectively. The fusion combines multi-scale information into a unified vessel representation, enabling richer vessel-aware refinement than a single-scale convolutional branch.

\subsection*{Feature fusion and final refinement}
After attention modulation and multi-scale context extraction, the resulting features are concatenated and fused using a $3 \times 3$ convolution, batch normalization, ReLU activation, and dropout (0.1). A subsequent $3 \times 3$ refinement block generates the final output feature map (Eq.~(\ref{eq20})):
\begin{equation}\label{eq20}
\begin{aligned}
& F_{\text {combined }}=\left[F_{\text {attended }}, C_{\text {fused }}\right] \in \mathbb{R}^{B \times 2 C \times H \times W}, \\
& F_{\text {fused }}=\operatorname{Dropout}_{0.1}\left(\delta\left(\mathrm{BN}\left(\operatorname{Conv}_{3 \times 3}^{2 C \rightarrow C}\left(F_{\text {combined }}\right)\right)\right)\right) \in \mathbb{R}^{B \times C \times H \times W}, \\
& F_{\text {refined }}=\delta\left(\mathrm{BN}\left(\operatorname{Conv}_{3 \times 3}^{C \rightarrow C}\left(F_{\text {fused }}\right)\right)\right) \in \mathbb{R}^{B \times C \times H \times W} .
\end{aligned}
\end{equation}
Overall, SGFA serves as a vessel-focused refinement. The refined features are mapped to pixel-wise vessel logits using a lightweight segmentation head comprising two $3 \times 3$ convolutions and a final $1 \times 1$ projection layer. This shallow design preserves subtle vessel structures while reducing over-smoothing of thin branches and vessel boundaries.

\subsection{Boundary-weighted adaptive compound loss (BWACL)}
Accurate coronary DSA vessel segmentation requires not only effective feature representation but also robust optimization objectives. Severe class imbalance and low-contrast vessel boundaries make thin-vessel preservation, accurate localization, and vessel continuity challenging. To address these issues, the proposed BWACL combines boundary-aware spatial weighting, weighted focal loss, weighted Dice loss, and adaptive loss balancing.

\subsubsection{Boundary weight generation}
The proposed loss emphasizes clinically important thin-vessel boundaries through a pixel-wise weight map $W \in \mathbb{R}^{B \times 1 \times H \times W}$, assigning higher weights (up to $1.5\times$) to thin-vessel boundary pixels and a baseline weight of $1.0$ elsewhere. Let $Y \in {0,1}^{B \times 1 \times H \times W}$ denote the ground-truth mask. The Euclidean distance transforms of the foreground and background regions, denoted by $D_{\mathrm{fg}}$ and $D_{\mathrm{bg}}$, represent the distances to the nearest vessel and background pixels, respectively. The local vessel radius is approximated by $D_{\mathrm{fg}}+D_{\mathrm{bg}}$, and thin vessels are identified by thresholding this radius map as follows:
\vspace{-3pt}
\begin{equation}\label{eq21}
\begin{aligned}
D_{\mathrm{fg}} = \operatorname{EDT}(Y),  
D_{\mathrm{bg}} =\operatorname{EDT}(1-Y), 
\mathbf{M}_{thin} =\mathbf{1}[(\mathrm{D}_{\mathrm{fg}}+\mathrm{D}_{\mathrm{bg}})<5]
\end{aligned}
\end{equation}
where, $\mathbf{1}$ denotes the indicator function and the threshold of $5$ pixels identifies thin vascular structures.To further emphasize vessel boundaries, we exploit the fact that boundary regions occur where both $D_{\text{fg}}$ and $D_{\text{bg}}$ are small. The boundary emphasis and final weight maps are computed as follows:
\vspace{-5pt}
\begin{equation}\label{eq24}
\mathbf{W}_{\text {boundary }}=\exp \left(-\frac{\left(\mathbf{D}_{f g}+\mathbf{D}_{b g}\right)^2}{2(\delta+\mathbf{R} \cdot \epsilon)^2}\right),
\mathbf{W}=\mathbf{1}+(\alpha-\mathbf{1}) \cdot \mathbf{W}_{\text {boundary }} \cdot \mathbf{M}_{\text {thin }}
\end{equation}
where, $\delta=1\times10^{-4}$ prevents division by zero, $\epsilon=0.01$ controls boundary width, and $\alpha=1.5$ controls the maximum boundary emphasis. Only thin-vessel boundary pixels ($M_{\text{thin}}=1$ and $W_{\text{boundary}}>0$) receive additional weight, while other pixels retain unit weight, emphasizing thin-vessel boundaries.

\subsubsection{Weighted focal loss}
Although focal loss \citep{lin2017focal} reduces the influence of easy examples and emphasizes hard pixels, it assigns equal importance across the image. In coronary DSA segmentation, thin-vessel boundaries are more challenging than vessel interiors and background regions. To address this limitation, the proposed boundary-aware weight map $W$ is incorporated into focal loss. Let $Z$ denote the network logits, $P=\sigma(Z)$ the predicted probabilities, and $P_{t,i}$ the predicted probability of the ground-truth for pixel $i$. The weighted focal loss is defined in (Eq.~(\ref{eq27})):
\vspace{-5pt}
\begin{equation}\label{eq27}
\mathcal{L}_{\mathrm{focal}}=\frac{1}{N} \sum_{i=1}^N \mathbf{W}_i \cdot\left(1-\mathbf{P}_{t, i}\right)^\gamma \cdot \mathcal{L}_{\mathrm{BCE}}\left(\mathbf{Z}_i, \mathbf{Y}_i\right)
\end{equation}
where, $N=B\times H\times W$ denotes the total number of pixels, $\mathcal{L}_{\text{BCE}}(Z_i,Y_i)$ is the binary cross-entropy loss for pixel $i$, and $\gamma \geq 0$ controls the focusing effect ($\gamma=1$ in all experiments). This formulation combines focal weighting for hard examples with spatial weighting for thin-vessel boundaries, guiding optimization toward uncertain predictions and clinically important vessel regions.

\subsubsection{Weighted dice loss}
Dice loss \citep{milletari2016v} is widely used in medical image segmentation because it directly measures overlap between prediction and ground truth. However, standard Dice treats all pixels equally and does not emphasize thin-vessel boundaries. Therefore, the proposed weight map $W$ is incorporated into the Dice formulation:
\vspace{-9pt}
\begin{equation}\label{eq29}
\mathcal{L}_{\text {Dice }}=1-\frac{2 \sum_{i=1}^N \mathbf{W}_i \cdot \mathbf{P}_i \cdot \mathbf{Y}_i+\delta}{\sum_{i=1}^N \mathbf{W}_i \cdot\left(\mathbf{P}_i+\mathbf{Y}_i\right)+\delta}
\end{equation}
where, $P_i$ denotes the predicted probability for pixel $i$, $Y_i$ the ground truth, $W_i$ the boundary weight, and $\delta$ a small constant ($\delta=1\times10^{-4}$) for numerical stability. By assigning larger weights to thin-vessel boundary pixels, the weighted Dice loss improves overlap in these regions and complements the hard-example emphasis of the weighted focal loss.

\subsubsection{Adaptive loss balancing}
Using fixed weights to combine the focal and Dice losses may be suboptimal because the dominant prediction errors vary during training. To address this issue, an adaptive balancing mechanism is introduced to dynamically adjust the contribution of the two loss terms according to the current recall and boundary errors.

First, recall error (missed vessels) counts the number of vessel pixels that the model incorrectly predicts as background (false negatives) as in (Eq.~(\ref{eq30})):
\vspace{-5pt}
\begin{equation}\label{eq30}
\mathcal{E}_{\text {recall }}=\sum_{i=1}^N \mathbf{Y}_i \cdot\left(1-1\left[\mathbf{P}_i>0.5\right]\right)
\end{equation}
where, $Y_i$ is $1$ for vessel pixels and $0$ for background. Thus, $\mathcal{E}_{\text{recall}}$ counts false negatives—vessel pixels that the model failed to detect. A larger value of $\mathcal{E}_{\text{recall}}$ indicates insufficient vessel recovery.

Second, boundary error (false positives on boundaries) counts the number of thin boundary pixels that the model incorrectly predicts as vessel, computed as in (Eq.~(\ref{eq31})):
\vspace{-5pt}
\begin{equation}\label{eq31}
\mathcal{E}_{\text {boundary }}=\sum_{i=1}^N 1\left[\mathbf{W}_i>1.0\right] \cdot 1\left[\mathbf{P}_i>0.5\right] \cdot\left(1-\mathbf{Y}_i\right)
\end{equation}
where $\mathbf{1}[W_i>1.0]$ selects thin-vessel boundary pixels, $\mathbf{1}[P_i>0.5]$ selects predicted vessel pixels, and $(1-Y_i)$ denotes background pixels. A high $\mathcal{E}_{\text{boundary}}$ indicates excessive vessel predictions along boundaries, leading to inaccurate edge localization. Based on these error measures, the adaptive balancing factor $\beta$ is defined in Eq.~(\ref{eq32}):
\vspace{-3pt}
\begin{equation}\label{eq32}
\beta=\sigma\left(\frac{\mathcal{E}_{\text {recall }}-\mathcal{E}_{\text {boundary }}}{1000}\right)
\end{equation}
where $\sigma$ denotes the sigmoid function, mapping the relative dominance of missed-vessel and boundary false-positive errors to $(0,1)$, and the scaling factor $1000$ stabilizes its response and prevents sharp changes. When $\mathcal{E}_{\text{recall}} > \mathcal{E}_{\text{boundary}}$, $\beta > 0.5$ increases the contribution of the focal term to encourage vessel recovery. Conversely, when $\mathcal{E}_{\text{boundary}} > \mathcal{E}_{\text{recall}}$, $\beta < 0.5$ increases the contribution of the Dice term to improve boundary refinement. When the two errors are balanced, $\beta$ approaches $0.5$. The final Boundary-Weighted Adaptive Compound Loss is computed as follows:
\vspace{-3pt}
\begin{equation}\label{eq33}
\mathcal{L}_{\text {total }}=\beta \cdot \mathcal{L}_{\text {focal }}+(1-\beta) \cdot \mathcal{L}_{\text {Dice }}
\end{equation}
This adaptive formulation dynamically balances vessel detection and boundary refinement according to the model's current errors. Overall, the proposed loss addresses severe class imbalance while improving thin-vessel continuity and boundary precision in challenging DSA images.

\section{Experiment settings}
\label{sec:experiment}

\subsection{Dataset }
The DSA dataset employed in this work was originally provided by the General Hospital of Southern Theater Command and previously described in~\citep{ahmed2026msa,pu2023semi}. It comprises 300 coronary X-ray angiographic sequences acquired from 50 patients, with each frame having a resolution of $512 \times 512$ pixels. The dataset includes separate imaging sessions for the right and left coronary arteries. From these sequences, a total of 150 right coronary artery images and 150 left coronary artery images were selected by clinical experts based on diagnostic relevance. All selected frames were subsequently resized to $256 \times 256$ pixels during preprocessing. For each coronary subset, 5-fold cross-validation was performed on all 150 images. In each fold, 120 images were used for training and the remaining 30 images were used for testing. Final performance was obtained by averaging the results across all five folds.

\subsection{Implementation details}
The proposed HTC-SGA Former was implemented using the PyTorch and PyTorch Lightning frameworks. All experiments were conducted on an NVIDIA RTX 4090 GPU with 24 GB of VRAM. Each coronary DSA image and its corresponding ground truth label were resized to a fixed resolution of $256 \times 256$ pixels and normalized to the range $[-1, 1]$ using a mean and standard deviation of 0.5. During training, we applied data augmentation techniques, including random brightness and contrast adjustments \citep{wu2025adaptive}, to improve model generalization. Optimization was performed using the Adam optimizer with an initial learning rate of $1 \times 10^{-4}$, decayed by a factor of 0.1 at 60\% and 80\% of the total 100 training epochs. A batch size of 5 was maintained throughout training.

\subsection{Evaluation metrics} 
Segmentation performance was evaluated using four standard metrics: Recall, Dice coefficient (DSC), average surface distance (ASD), and average contour distance (ACD). DSC and Recall measure region overlap and vessel recovery, whereas ASD and ACD assess boundary and contour accuracy. Detailed definitions and mathematical formulations of these metrics are provided in \citep{ahmed2026msa}.
 
\section{Results and discussion}
This section presents the experimental evaluation of HTC-SGA Former and the proposed BWACL for coronary DSA vessel segmentation. We first assess the effectiveness of BWACL independently of the proposed architecture by applying it to four widely used segmentation backbones: U-Net, Attention U-Net, U-Net++, and Attention U-Net++. We then compare HTC-SGA Former with 14 state-of-the-art segmentation methods to evaluate its segmentation accuracy and parameter efficiency. Since the dataset contains separate right and left coronary artery subsets with distinct anatomical characteristics, results are reported separately for each subset. Performance is evaluated using Recall, Dice, ASD, and ACD under 5-fold cross-validation, with results reported as mean values.

\begin{table*}[htbp]
\centering
\caption{Performance comparison of BCE+Dice and BWACL on four backbone architectures. Results are reported as mean values over 5-fold cross-validation on the right and left coronary DSA subsets.}
\label{table1}
\renewcommand{\arraystretch}{1.1}
  \setlength{\tabcolsep}{1.8pt}
   \resizebox{\linewidth}{!}{%
\begin{tabular}{llcccccccc}
\toprule
\multirow{2}{*}{\textbf{Architecture}} &
\multirow{2}{*}{\textbf{Loss}} &
\multicolumn{4}{c}{\textbf{Right Coronary Subset}} &
\multicolumn{4}{c}{\textbf{Left Coronary Subset}} \\
\cmidrule(lr){3-6} \cmidrule(lr){7-10}
& &
\textbf{Recall$\uparrow$} &
\textbf{Dice$\uparrow$} &
\textbf{ASD$\downarrow$} &
\textbf{ACD$\downarrow$} &
\textbf{Recall$\uparrow$} &
\textbf{Dice$\uparrow$} &
\textbf{ASD$\downarrow$} &
\textbf{ACD$\downarrow$} \\
\midrule
\multirow{2}{*}{\textbf{U-Net \citep{ronneberger2015u}}}
& Bce+Dice
& 85.79 & 87.39 & 0.8170 & 0.7955
& 81.17 & 81.82 & 1.1161 & 1.0870 \\
& BWACL
& \textbf{86.43} & \textbf{87.64} & \textbf{0.7598} & \textbf{0.7430}
& \textbf{85.06} & \textbf{82.25} & \textbf{1.0669} & \textbf{1.0579} \\
\midrule
\multirow{2}{*}{\textbf{AttU-Net \citep{oktay2018attention}}}
& Bce+Dice
& 85.89 & 87.41 & 0.7961 & 0.7782
& 81.90 & 82.28 & 1.1120 & 1.0837 \\
& BWACL
& \textbf{86.10} & \textbf{87.71} & \textbf{0.7778} & \textbf{0.7587}
& \textbf{83.36} & \textbf{82.59} & \textbf{1.0678} & \textbf{1.0486} \\
\midrule
\multirow{2}{*}{\textbf{U-Net++ \citep{Zongwei_Rahman_Nima_Jianming_2022}}}
& Bce+Dice
& 84.15 & 86.34 & 0.8259 & 0.8075
& 80.31 & 80.81 & 1.1074 & 1.0882 \\
& BWACL
& \textbf{86.18} & \textbf{87.04} & \textbf{0.7715} & \textbf{0.7585}
& \textbf{82.82} & \textbf{82.18} & \textbf{1.0841} & \textbf{1.0682} \\
\midrule
\multirow{2}{*}{\textbf{AttU-Net++ \citep{li2020attention}}}
& Bce+Dice
& 85.19 & 86.85 & 0.8108 & 0.7959
& 80.87 & 81.33 & 1.1143 & 1.0916 \\
& BWACL
& \textbf{85.27} & \textbf{87.54} & \textbf{0.7914} & \textbf{0.7687}
& \textbf{83.90} & \textbf{82.23} & \textbf{1.0893} & \textbf{1.0740} \\
\bottomrule
\end{tabular}
}
\vspace{0.05cm}

\begin{minipage}{\textwidth}
\small
\textbf{Notes:}
AttU-Net --- Attention U-Net and AttU-Net++ --- Attention U-Net++.
\end{minipage}
% \vspace{-10pt}
\end{table*}

\subsection{Effectiveness of the boundary-weighted adaptive compound loss (BWACL)}
To evaluate the contribution of BWACL, each backbone was trained under identical settings using either the baseline (BCE+Dice) loss or the proposed BWACL. As shown in Table~\ref{table1}, replacing (BCE+Dice) with BWACL consistently improves segmentation performance across all four architectures on both coronary artery subsets. On the right subset, U-Net Recall and Dice increase from 85.79/87.39 to 86.43/87.64, while ASD and ACD decrease from 0.8170/0.7955 to 0.7598/0.7430, respectively. Similar improvements are observed for Attention U-Net, U-Net++, and Attention U-Net++, as well as across the left subset. Overall, these results demonstrate that BWACL consistently improves segmentation accuracy and boundary delineation across different architectures and coronary artery subsets.

The results in Table~\ref{table1} show that BWACL provides a stronger optimization objective than the conventional (BCE+Dice) loss for coronary DSA vessel segmentation. Improved Recall indicates better recovery of thin and weak distal vessels under severe class imbalance, while the reductions in ASD and ACD confirm enhanced boundary localization through the proposed boundary-aware weighting strategy. Although the left coronary subset is more challenging due to its complex vascular structure and finer distal branches, BWACL consistently improves performance across all evaluated models. These findings demonstrate the effectiveness of BWACL for vessel recovery and boundary localization across different backbone architectures.

\begin{figure}
    \centering
    \begin{minipage}{0.47\textwidth}
        \centering
        \includegraphics[width=\textwidth]{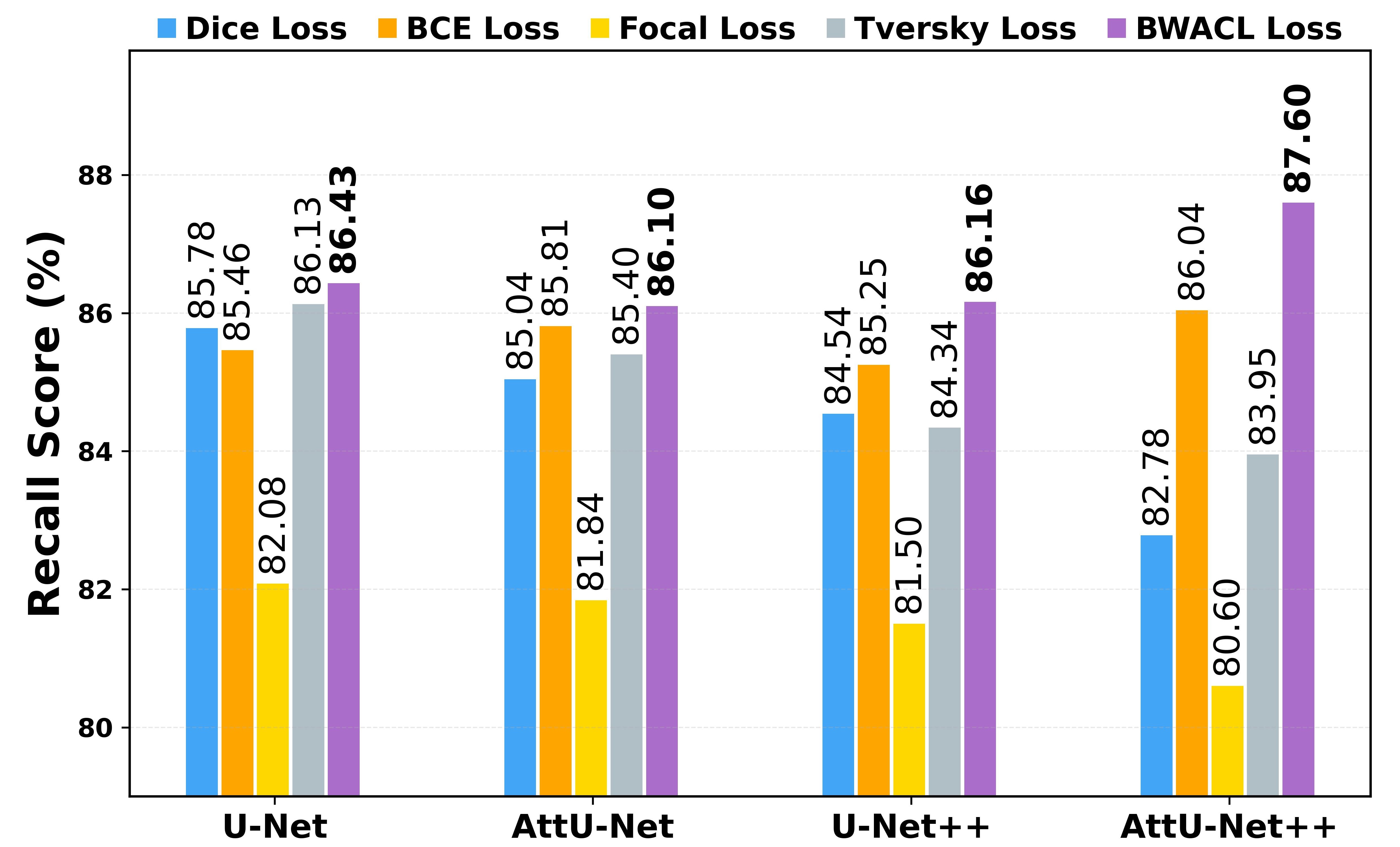}
        \subcaption{}
        \vspace{-5pt} 
        \label{fig6a}
    \end{minipage}%
    \begin{minipage}{0.47\textwidth}
        \centering
        \includegraphics[width=\textwidth]{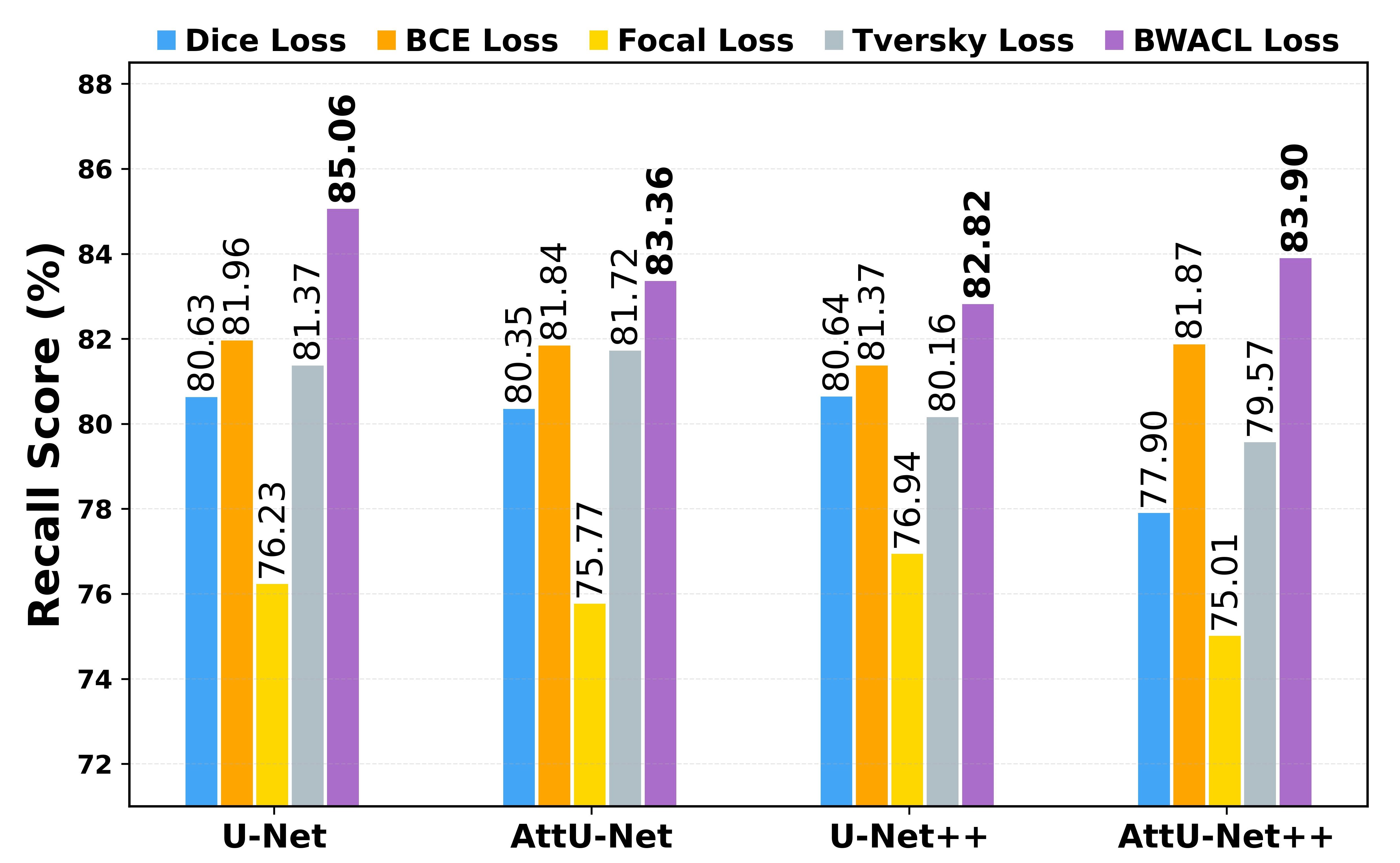}
        \subcaption{}
        \vspace{-5pt} 
        \label{fig6b}
    \end{minipage}
    \caption{Recall comparison of different loss functions across U-Net, Attention U-Net, U-Net++, and Attention U-Net++. (a) Right coronary artery subset. (b) Left coronary artery subset. BWACL achieves the highest recall.}
    \label{fig6}
    \vspace{-8pt} 
\end{figure}

To further validate the vessel-recovery capability of BWACL, Fig.~\ref{fig6} compares its Recall performance against Dice, BCE, Focal, and Tversky losses across the four backbone architectures on the right and left coronary artery subsets. As shown in Fig.~\ref{fig6}a and Fig.~\ref{fig6}b, BWACL consistently achieves the highest Recall across all evaluated architectures and subsets. On the right subset, Recall reaches up to $87.60\%$, while on the more challenging left subset it reaches $85.06\%$, outperforming all competing losses. 
The consistent Recall improvement in Fig.~\ref{fig6} indicates that BWACL more effectively reduces false negatives than all competing losses, particularly for thin and low-contrast distal vessels, resulting in improved vessel recovery across all evaluated backbones and coronary subsets. These results demonstrate that emphasizing thin-vessel boundaries while adaptively balancing vessel recovery and boundary refinement provides a more effective optimization objective for coronary vessel segmentation.

\subsection*{Qualitative analysis of BWACL}
\begin{figure}
	\centering
	\includegraphics[width=0.8\textwidth]{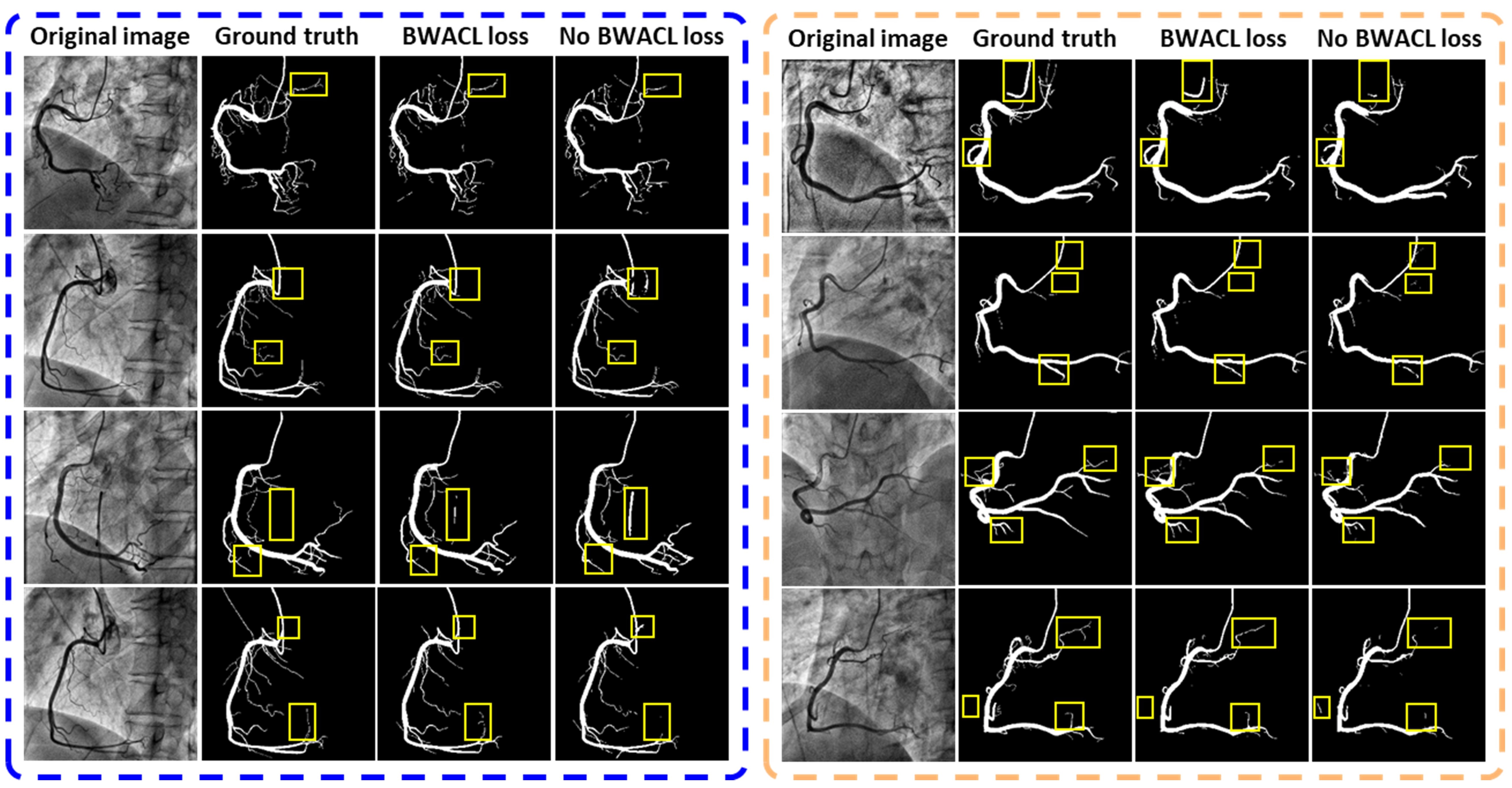}
	\caption{Qualitative segmentation comparison on the right coronary subset using U-Net (blue border) and U-Net++ (orange border) trained with and without BWACL. Yellow rectangles highlight regions where BWACL improves thin-vessel recovery, boundary delineation, and false-positive suppression.}
	\label{fig7}
	\vspace{-11pt} 
\end{figure}

\begin{figure}
	\centering
	\includegraphics[width=0.8\textwidth]{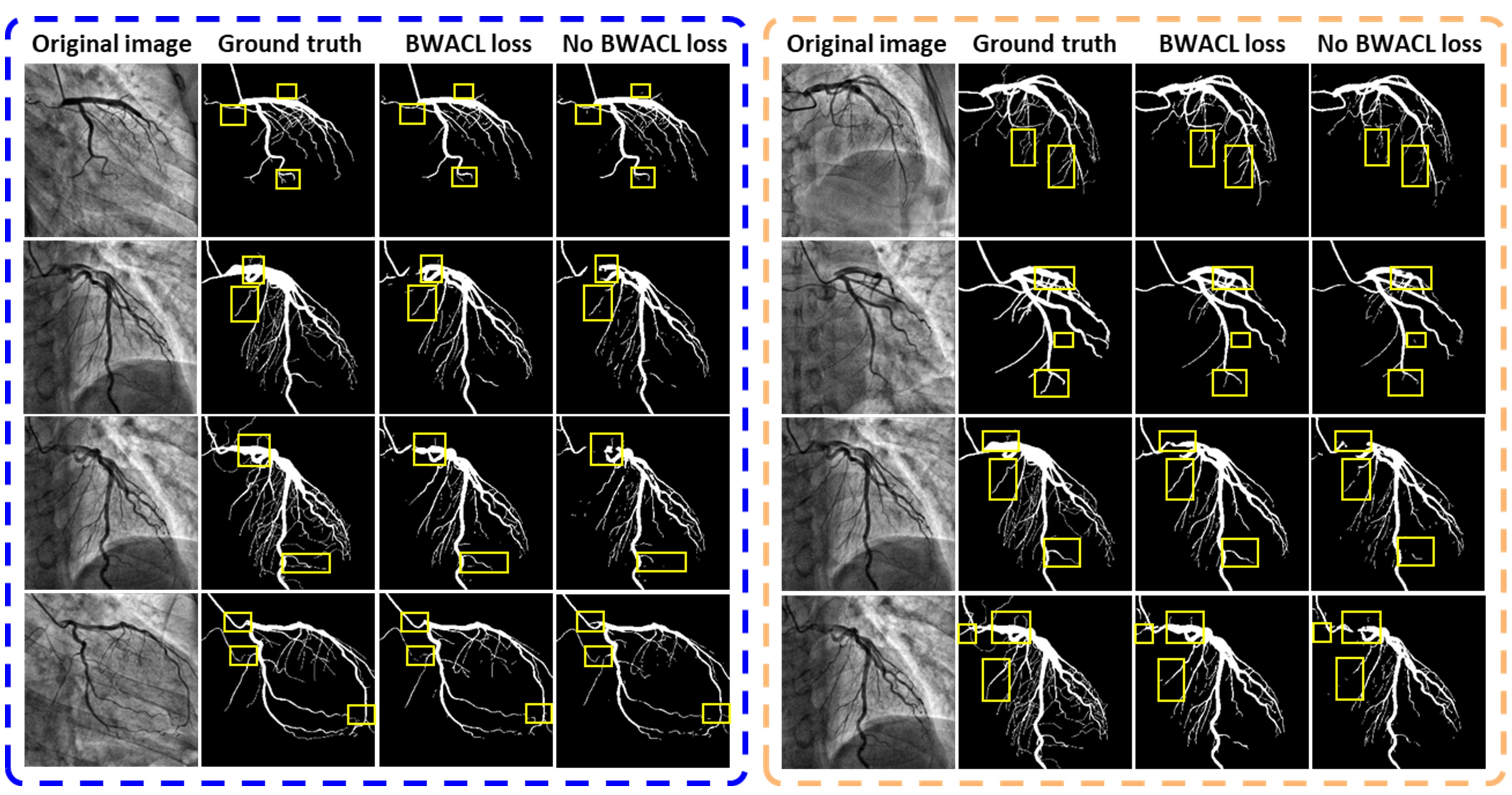}
	\caption{Qualitative segmentation comparison on the left coronary subset using Attention U-Net (blue border) and Attention U-Net++ (orange border) trained with and without BWACL. Yellow rectangles highlight regions where BWACL improves thin-vessel recovery, boundary delineation, and false-positive suppression.}
	\label{fig8}
	\vspace{-11pt} 
\end{figure}

Beyond the quantitative results, Figs.~\ref{fig7} and \ref{fig8} present qualitative comparisons on the right and left coronary artery subsets, respectively. Results are shown for U-Net and U-Net++ on the right subset, and Attention U-Net and Attention U-Net++ on the left subset. Highlighted regions indicate challenging areas containing thin vessels, weak responses, boundary ambiguity, fragmented structures, and false-positive interference.

Models trained without BWACL often exhibit incomplete thin-vessel recovery, fragmented vessels, and increased false-positive responses in surrounding anatomy. In contrast, incorporating BWACL improves vessel continuity, preserves weak distal branches, reduces fragmented predictions, and enhances boundary delineation across all evaluated backbones. These improvements are particularly evident in low-contrast thin-vessel and bifurcation regions, where severe class imbalance bias conventional losses toward background prediction. These qualitative observations are consistent with the proposed boundary-aware and adaptive optimization strategy and further support the quantitative improvements reported in Table~\ref{table1}, and Fig.~\ref{fig6}.

\subsection{Performance comparison with state-of-the-art methods}
After independently validating BWACL, we evaluate the complete HTC-SGA Former framework to determine whether the proposed encoder-decoder architecture combined with BWACL loss further improves coronary DSA vessel segmentation. The proposed method is compared with 14 state-of-the-art CNN-, attention-, and Transformer-based segmentation models on the right and left coronary DSA subsets, with results reported in Table~\ref{table2} as mean values over 5-fold cross-validation. In addition, a one-tailed t-test is performed to assess the statistical significance of the improvements achieved by HTC-SGA Former. 

As shown in Table~\ref{table2}, HTC-SGA Former achieves the best performance on both coronary DSA subsets across all evaluation metrics, while maintaining a lightweight architecture of only $0.81$M parameters. On the right subset, it attains $88.58$ Recall, $88.01$ Dice, $0.6944$ ASD, and $0.6882$ ACD, outperforming the strongest competitors, including MSA-UNet3+, which achieves ($86.29$ Recall, $87.64$ Dice, $0.7872$ ASD, and $0.7675$ ACD) and FR\_UNet reaching ($85.96$ Recall, $87.24$ Dice). Among Transformer-based methods, H2Former obtains $83.41$ Recall and $85.58$ Dice, while SwinUNet, MISSFormer, and Isunetv1 show higher ASD and ACD values, indicating weaker boundary and structural accuracy. Similarly, on the left subset, HTC-SGA Former achieves $85.23$ Recall, $82.89$ Dice, $1.0256$ ASD, and $1.0154$ ACD, surpassing the nearest competitors, including MSA-UNet3+ ($81.67$ Recall, $82.47$ Dice, $1.0822$ ASD, and $1.0565$ ACD) and IMFF\_Net reaches ($81.33$ Recall, $82.24$ Dice). These results demonstrate superior segmentation accuracy and boundary localization while maintaining a favorable balance between performance and model complexity.

The observed performance gains demonstrate that HTC-SGA Former consistently outperforms state-of-the-art segmentation models across both coronary subsets. Such improvements are particularly important in coronary DSA segmentation, where weak distal branches, low-contrast vessels, and overlapping background structures make vessel recovery and boundary delineation highly challenging. Higher Recall and Dice values indicate improved preservation of thin and weak vessels, while lower ASD and ACD reflect more accurate boundary localization. These gains are consistent with the proposed combination of CNN-based feature extraction, MS-GLWA contextual modeling, SGFA refinement, and BWACL optimization. Notably, HTC-SGA Former also surpasses the strongest competing model, MSA-UNet3+, while requiring substantially fewer parameters. Despite using only $0.81$M parameters, HTC-SGA Former achieves a favorable balance between segmentation accuracy and computational efficiency.

\begin{figure}
    \centering
    \begin{minipage}{0.47\textwidth}
        \centering
        \includegraphics[width=\textwidth]{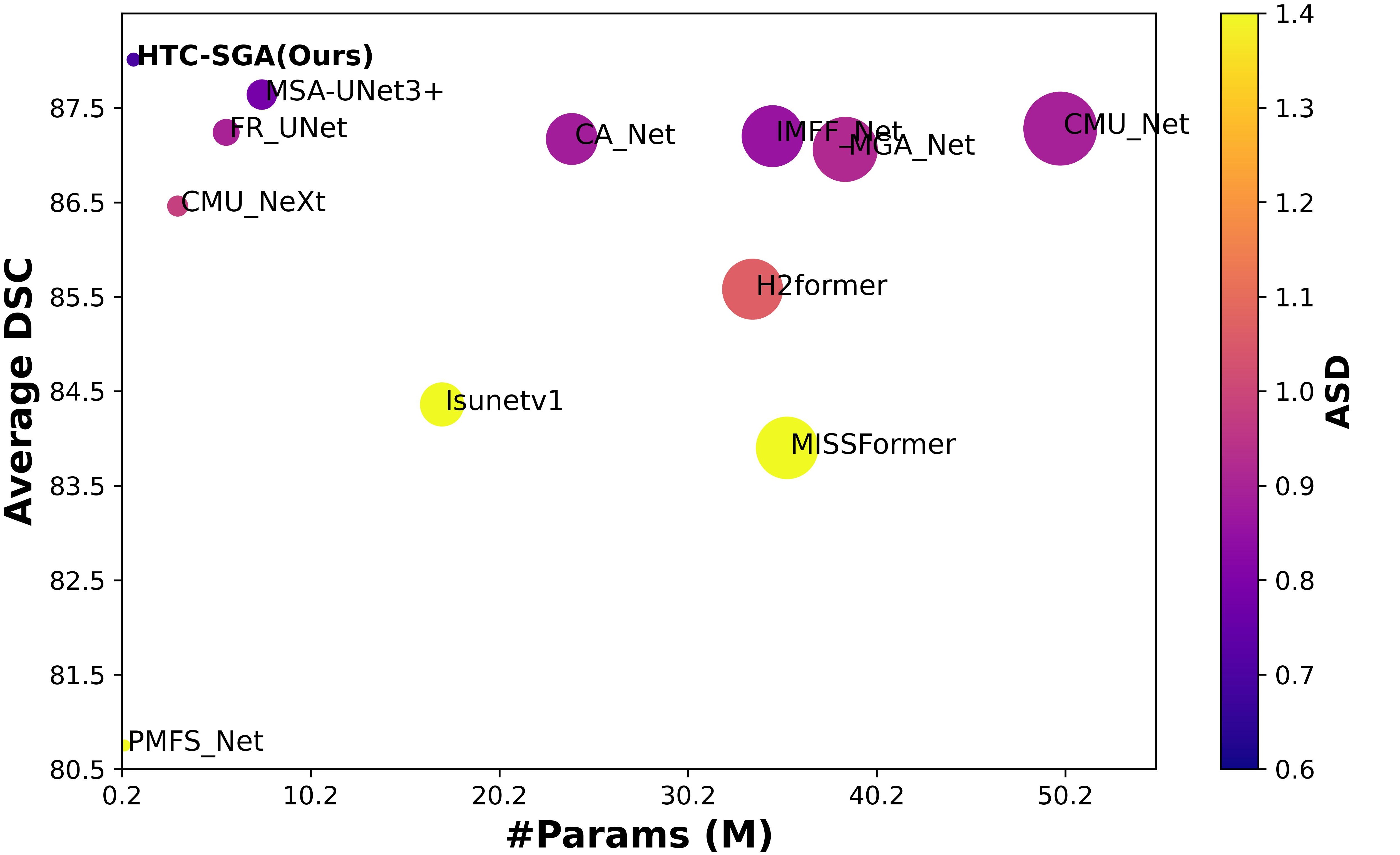}
        \subcaption{}
        \label{fig11a}
    \end{minipage}%
    \begin{minipage}{0.47\textwidth}
        \centering
        \includegraphics[width=\textwidth]{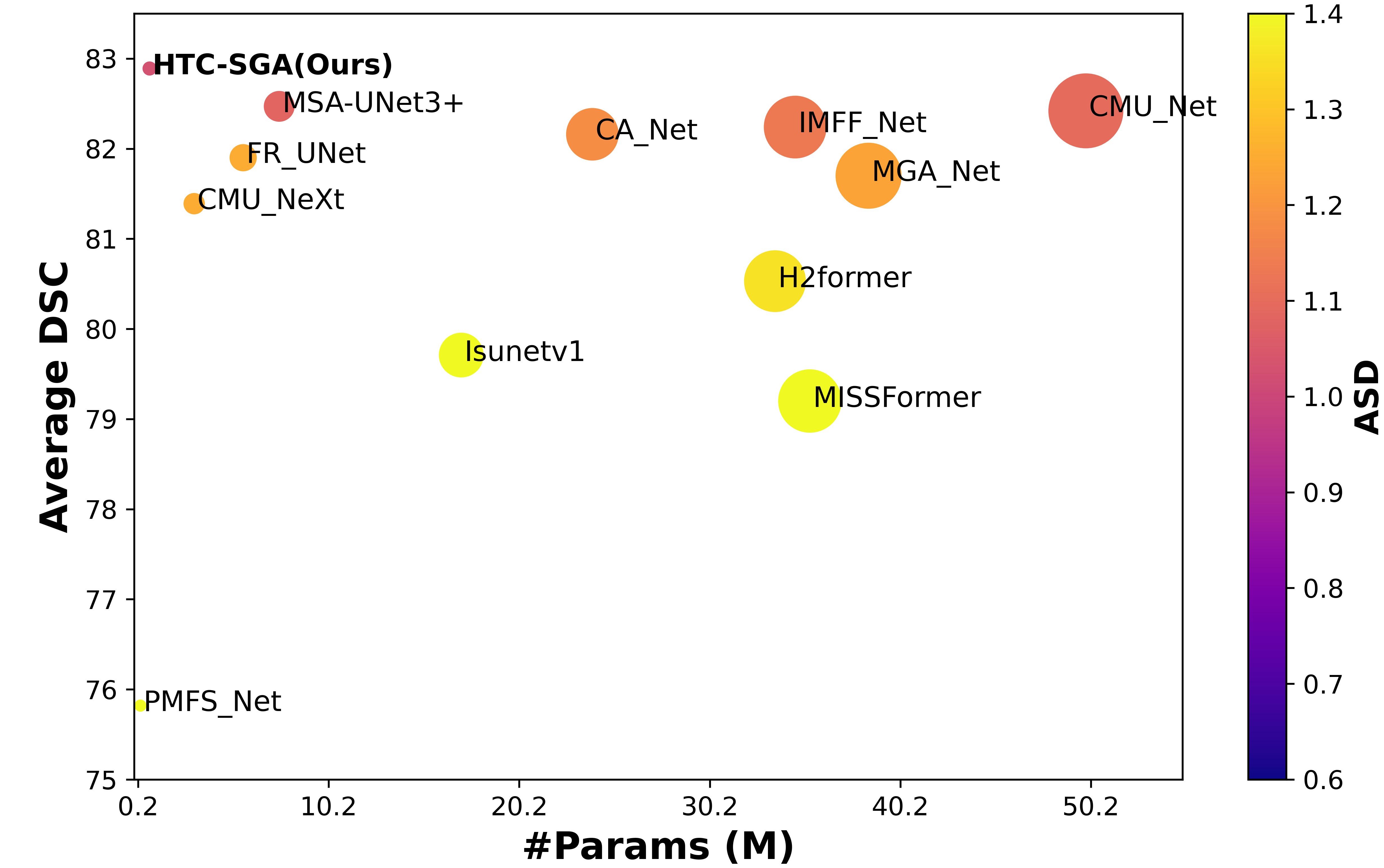}
        \subcaption{}
        \label{fig11b}
    \end{minipage}
    \caption{The figure illustrates the trade-off between parameter count (M), Dice coefficient (DSC), and ASD for different segmentation models. Point size and color represent model complexity and ASD, respectively. (a) Right coronary artery subset. (b) Left coronary artery subset.}
    \label{fig11}
    \vspace{-8pt} 
\end{figure}

Fig.~\ref{fig11}a and Fig.~\ref{fig11}b compare the trade-off between model efficiency and segmentation performance on the right and left subsets, respectively. The horizontal axis represents parameter count, while the vertical axis shows Dice coefficient (DSC). Point size and color indicate model complexity and ASD, respectively, where darker colors denote lower ASD. Across both subsets, HTC-SGA Former achieves a favorable accuracy-efficiency trade-off, attaining the highest DSC and lowest ASD while maintaining a compact architecture. Although PMFS\_Net is more lightweight, its lower DSC and higher ASD indicate reduced segmentation accuracy, whereas several larger CNN- and Transformer-based models require substantially more parameters while producing inferior results. These findings highlight the strong parameter efficiency and boundary localization capability of HTC-SGA Former.

\begin{table*}[htbp]
\centering
\caption{Performance comparison of state-of-the-art segmentation models on the right and left coronary DSA subsets. Results are reported as mean values over 5-fold cross-validation. Statistical significance is assessed using a one-tailed $t$-test.}
\setlength{\tabcolsep}{4pt}
\renewcommand{\arraystretch}{1.1}
\setlength{\tabcolsep}{3.0pt}
\resizebox{\linewidth}{!}{%
\begin{tabular}{lccccccccc}
\toprule
\multirow{2}{*}{\textbf{Architecture}} &
\multicolumn{4}{c}{\textbf{Right Coronary Subset}} &
\multicolumn{4}{c}{\textbf{Left Coronary Subset}} &
\multirow{2}{*}{\textbf{Param (M)}} \\
\cmidrule(lr){2-5} \cmidrule(lr){6-9}
& \textbf{Recall $\uparrow$} & \textbf{Dice $\uparrow$} & \textbf{ASD $\downarrow$} & \textbf{ACD $\downarrow$}
& \textbf{Recall $\uparrow$} & \textbf{Dice $\uparrow$} & \textbf{ASD $\downarrow$} & \textbf{ACD $\downarrow$}
& \\
\midrule

CMU\_Net \citep{tang2023cmu} 
& 85.45$\star\star$ & 87.28$\diamond\diamond$ & 0.8919$\star\star$ & 0.8720$\star\star$
& 81.27$\star\star$ & 82.42$\diamond$ & 1.0970$\diamond$ & 1.0728$\diamond$
& 49.93 \\

FR\_UNet \citep{liu2022full} 
& 85.96$\star\star$ & 87.24$\star$ & 0.8972$\star\star$ & 0.8920$\star\star$
& \underline{81.88}$\star\star$ & 81.90$\star\star$ & 1.2502$\star\star$ & 1.2200$\star\star$
& 5.72 \\

CA\_Net \citep{xie2023canet} 
& 85.12$\star\star$ & 87.17$\star$ & 0.8832$\diamond\diamond$ & 0.8542$\star\star$
& 81.68$\star\star$ & 82.16$\diamond$ & 1.1834$\star\star$ & 1.1528$\diamond\diamond$
& 24.04 \\

CMU\_Next \citep{tang2024cmunext} 
& 82.47$\star\star$ & 86.46$\diamond\diamond$ & 0.9812$\star\star$ & 0.9439$\star\star$
& 79.09$\star\star$ & 81.39$\star\star$ & 1.2500$\star\star$ & 1.2030$\star\star$
& 3.15 \\

BCU\_Net \citep{zhang2023bcu} 
& 84.87$\star\star$ & 86.77$\diamond\diamond$ & 0.9561$\star\star$ & 0.9304$\star\star$
& 79.79$\star\star$ & 81.51$\diamond\diamond$ & 1.2025$\star\star$ & 1.1719$\diamond\diamond$
& 102.23 \\

MGA\_Net \citep{gao2023multi}  
& 85.35$\star\star$ & 87.06$\star$ & 0.9170$\diamond\diamond$ & 0.8905$\diamond\diamond$
& 80.61$\star\star$ & 81.70$\diamond\diamond$ & 1.2321$\star\star$ & 1.1912$\star\star$
& 38.53 \\

IMFF\_Net \citep{liu2024imff} 
& 84.13$\diamond\diamond$ & 87.20$\diamond$ & 0.858$\star\star$ & 0.8296$\star\star$
& 81.33$\diamond\diamond$ & 82.24$\diamond$ & 1.1343$\star$ & 1.1077$\star$
& 34.68 \\

PMFS\_Net \citep{zhong2025pmfsnet} 
& 80.26$\star\star$ & 80.75$\star\star$ & 1.4454$\star\star$ & 1.4084$\star\star$
& 74.65$\star\star$ & 75.82$\star\star$ & 1.6337$\star\star$ & 1.5716$\star\star$
& 0.33 \\

MSA-UNet3+ \citep{ahmed2026msa}
& \underline{86.29}$\star\star$	& \underline{87.64}$\diamond$ & \underline{0.7872}$\star\star$	& \underline{0.7675}$\star\star$ & 81.67$\star\star$	& \underline{82.47}$\star$ & \underline{1.0822}$\diamond\diamond$ & \underline{1.0565}$\star$ & 7.61 \\

Swinunet \citep{cao2022swin} 
& 79.55$\star\star$ & 81.79$\star\star$ & 1.8547$\star\star$ & 1.8323$\star\star$
& 76.40$\star\star$ & 77.30$\star\star$ & 1.8642$\star\star$ & 1.8295$\star\star$
& 85.9 \\

MISSFormer \citep{huang2022missformer}
& 81.74$\star\star$ & 83.90$\star\star$ & 1.4806$\star\star$ & 1.4384$\star\star$
& 78.04$\star\star$ & 79.20$\star\star$ & 1.5866$\star\star$ & 1.5345$\star\star$
& 35.45 \\

Isunetv1 \citep{pu2023semi}
& 81.69$\star\star$ & 84.36$\star\star$ & 1.4003$\star\star$ & 1.3597$\star\star$
& 78.58$\star\star$ & 79.71$\star\star$ & 1.5169$\star\star$ & 1.4698$\star\star$
& 17.16 \\

H2former \citep{he2023h2former}
& 83.41$\star\star$ & 85.58$\star\star$ & 1.0687$\star\star$ & 1.0155$\star\star$
& 79.59$\star\star$ & 80.53$\star\star$ & 1.3593$\star\star$ & 1.2956$\star\star$
& 33.62 \\

\textbf{HTC-SGA(Ours)}
& \textbf{88.58} & \textbf{88.01} & \textbf{0.6944} & \textbf{0.6882}
& \textbf{85.23} & \textbf{82.89} & \textbf{1.0256} & \textbf{1.0154}
& \textbf{0.81} \\

\bottomrule
\end{tabular}}

\vspace{0.05cm}
\begin{minipage}{\textwidth}
\small
\textbf{Notes:} Bold values indicate the best result for each metric, while underlined values denote the second-best result. $\star\star p$-value $< 0.0005$, $\diamond\diamond p$-value $< 0.001$,
$\star p$-value $< 0.01$, $\diamond p$-value $< 0.05$, compared with HTC-SGA Former (ours).
\end{minipage}
\vspace{-8pt}
\label{table2}
\end{table*}

\begin{figure}
	\centering
	\includegraphics[width=0.90\textwidth]{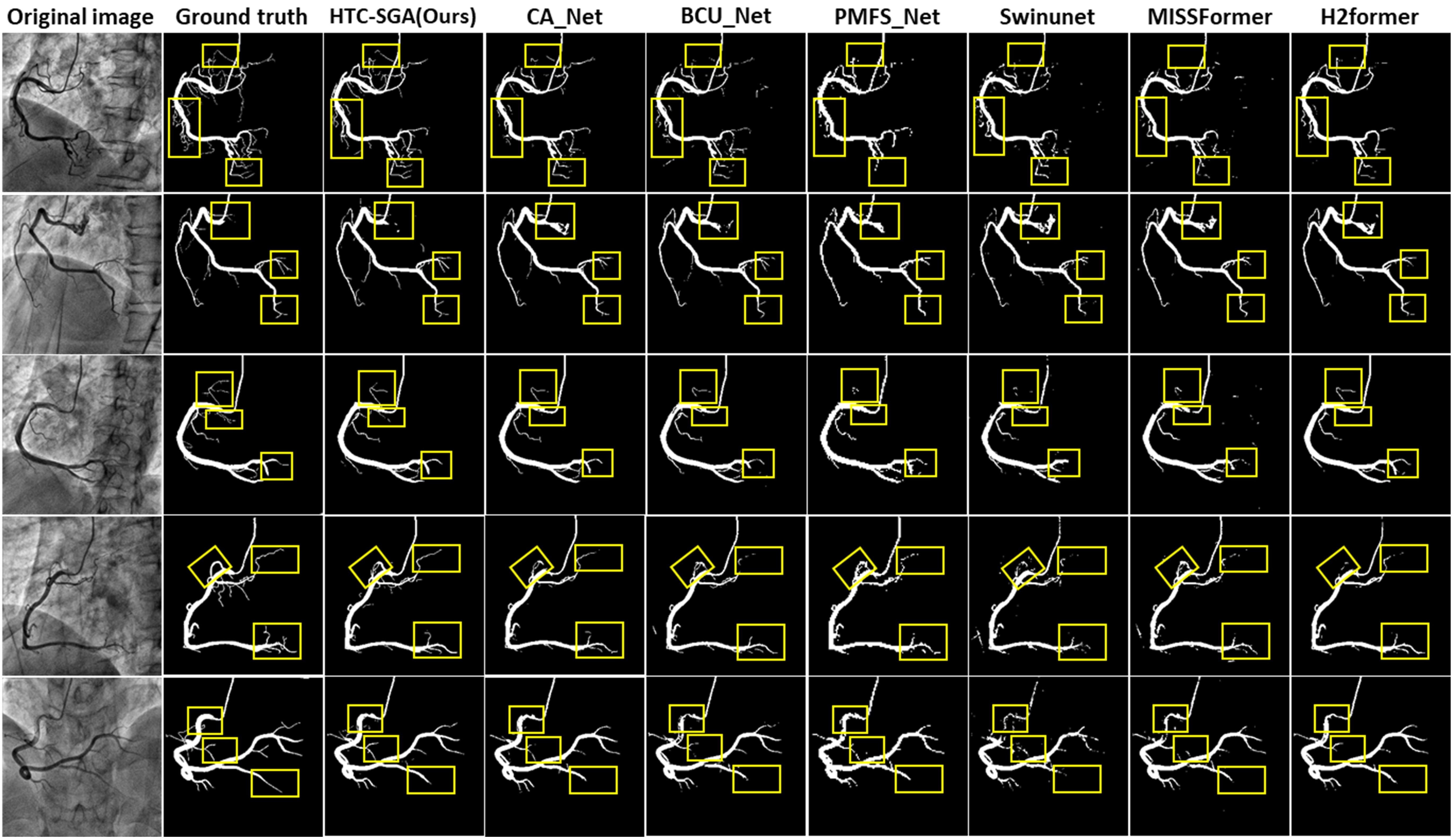}
	\caption{Qualitative comparison on the right coronary artery subset between HTC-SGA Former and state-of-the-art segmentation methods. Yellow rectangles highlight challenging regions with thin distal branches, vessel discontinuities, boundary ambiguity, and false-positive responses.}
	\label{fig9}
	\vspace{-11pt} 
\end{figure}

\begin{figure}
	\centering
	\includegraphics[width=0.90\textwidth]{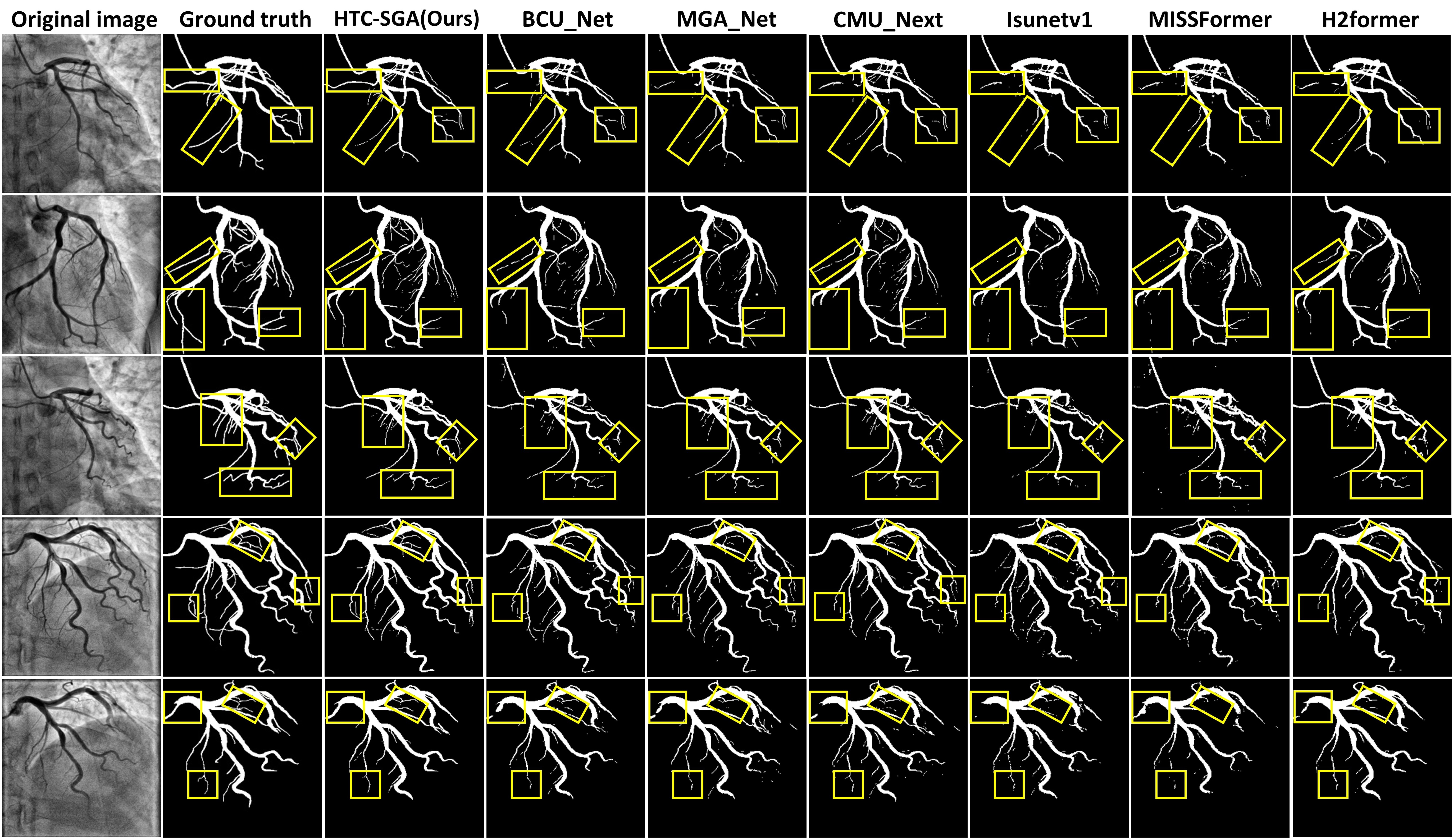}
	\caption{Qualitative comparison on the left coronary artery subset between HTC-SGA Former and state-of-the-art segmentation methods. Yellow rectangles highlight challenging regions with thin distal branches, complex bifurcations, boundary localization, and false-positive responses.}
	\label{fig10}
	\vspace{-11pt} 
\end{figure}

\subsection*{Qualitative comparison with state-of-the-art methods}
To further evaluate HTC-SGA Former qualitatively, Figs.~\ref{fig9} and \ref{fig10} present visualization comparisons on the right and left coronary artery subsets, respectively. Each figure compares HTC-SGA Former with CNN-based, hybrid CNN–Transformer, and Transformer-based models. Highlighted yellow regions indicate challenging areas containing thin vessels, weak distal branches, bifurcations, boundary ambiguity, and background interference. The qualitative results show that several existing methods suffer from incomplete thin-vessel recovery, fragmented continuity, blurred boundaries, and false-positive responses. CNN-based models generally preserve local vessel structures but often fail to maintain long-range continuity, leading to broken distal branches and missed weak vessels. In contrast, Transformer-based and hybrid CNN–Transformer models provide stronger contextual modeling but may produce over-smoothed predictions and inaccurate thin-vessel boundaries. In several highlighted regions, competing methods either miss weak vessel branches or generate discontinuous predictions that deviate from the ground truth.

Compared with competing methods, HTC-SGA Former produces segmentation results closer to the ground truth across both coronary subsets. The framework demonstrates superior preservation of thin distal branches, improved vascular continuity, and more accurate boundary delineation while suppressing false-positive responses. These improvements are particularly evident in low-contrast and complex bifurcation regions, where accurate modeling of local vessel morphology and broader anatomical context is essential. The qualitative results further validate the effectiveness of MS-GLWA, SGFA, and BWACL and are consistent with the quantitative comparisons in Tables~\ref{table2}.

\subsection{Ablation study}
To further validate HTC-SGA Former, an ablation study was conducted to evaluate the contributions of MS-GLWA module, SGFA module, and BWACL. Since MS-GLWA consists of a local branch (LB), global branch (GB), and enhanced multi-scale QKV generation (E-QKV) mechanism, the study also evaluates the contribution of these subcomponents. Results on the right and left coronary artery subsets are reported in Table~\ref{table5}.

\begin{table*}[htbp]
  \centering
  \caption{Ablation study of HTC-SGA Former on the right and left coronary DSA subsets, showing six configurations for evaluating the contributions of MS-GLWA (LB, GB, and E-QKV), SGFA, and BWACL. Symbols denote (\ding{51}) presence and (\ding{55}) absence of each component.}
  \setlength{\tabcolsep}{5pt}
  \renewcommand{\arraystretch}{1.1}
  % \begin{tabular}{@{}c@{\hspace{4pt}}ccccc@{\hspace{8pt}}cccc@{\hspace{8pt}}cccc@{}}
  \begin{tabular}{@{}c@{\hspace{4pt}}ccccc@{\hspace{8pt}}cccc@{\hspace{16pt}}cccc@{}}
    \toprule
    \multicolumn{6}{c}{\textbf{Configuration}} &
    \multicolumn{4}{c}{\textbf{Right Coronary Subset}} &
    \multicolumn{4}{c}{\textbf{Left Coronary Subset}} \\
    % \cmidrule(lr){1-6} \cmidrule(lr){7-10} \cmidrule(lr){11-14}
    \cmidrule(lr){1-6} \cmidrule(l{4pt}r{4pt}){7-10} \cmidrule(l{4pt}r{4pt}){11-14}
    & \multicolumn{3}{c}{\textbf{MS-GLWA}} & \textbf{SGFA} & \textbf{BWACL}
    \\
    \cmidrule(lr){2-4} \cmidrule(lr){5-5} \cmidrule(lr){6-6}
    \textbf{Idx} & \textbf{LB} & \textbf{GB} & \textbf{E-QKV} & & &
    \textbf{Recall $\uparrow$} & \textbf{Dice $\uparrow$} & \textbf{ASD $\downarrow$} & \textbf{ACD $\downarrow$}
    & \textbf{Recall $\uparrow$} & \textbf{Dice $\uparrow$} & \textbf{ASD $\downarrow$} & \textbf{ACD $\downarrow$} \\ [0.1cm]
    \midrule
    1 & \ding{55} & \ding{55} & \ding{55} & \ding{55} & \ding{55}
    & 85.33 & 87.14 & 0.8280 & 0.8098
    & 81.55 & 82.45 & 1.113 & 1.085 \\ [0.1cm]
    2 & \ding{51} & \ding{55} & \ding{55} & \ding{55} & \ding{55}
    & 85.86 & 87.23 & 0.8391 & 0.8215
    & 82.42 & 82.54 & 1.095 & 1.072 \\ [0.1cm]
    3 & \ding{51} & \ding{51} & \ding{55} & \ding{55} & \ding{55}
    & 86.19 & 87.36 & 0.8294 & 0.8115
    & 82.73 & 82.66 & 1.089 & 1.064 \\ [0.1cm]
    4 & \ding{51} & \ding{51} & \ding{51} & \ding{55} & \ding{55}
    & 86.58 & 87.35 & 0.8432 & 0.8240
    & 83.60 & 82.58 & 1.066 & 1.050 \\ [0.1cm]
    5 & \ding{51} & \ding{51} & \ding{51} & \ding{51} & \ding{55}
    & 87.02 & 87.70 & 0.7670 & 0.7547
    & 83.93 & 82.88 & 1.032 & 1.020 \\ [0.1cm]
    6 & \ding{51} & \ding{51} & \ding{51} & \ding{51} & \ding{51}
    & \textbf{88.58} & \textbf{88.01} & \textbf{0.6944} & \textbf{0.6882}
    & \textbf{85.32} & \textbf{82.89} & \textbf{1.026} & \textbf{1.015} \\
    \bottomrule
  \end{tabular}%
  \label{table5}%

  \vspace{0.1cm}
  \begin{minipage}{\textwidth}
    \small
    \textbf{Notes:}
    LB \text{---} Local Branch, GB \text{---} Global Branch, \text{and} E-QKV \text{---} Enhanced Multi-Scale QKV.
  \end{minipage}
  \vspace{-11pt}
\end{table*}

\subsubsection{Effect of the multi-scale global-local window attention (MS-GLWA) module}
As shown in Table~\ref{table5}, progressively constructing MS-GLWA yields consistent gains on both subsets. Relative to the baseline without LB, GB, E-QKV, SGFA, and BWACL (Idx 1), adding the local branch alone (Idx 2) improves Recall from 85.33 to 85.86 on the right subset and from 81.55 to 82.42 on the left subset, while Dice increases from 87.14 to 87.23 and from 82.45 to 82.54, respectively. Adding the global branch (Idx 3) further raises Recall to 86.19 and 82.73, and Dice to 87.36 and 82.66 on the right and left subsets, respectively. When E-QKV is included (Idx 4), Recall further improves to 86.58 on the right subset and 83.60 on the left subset.
These results show that MS-GLWA benefits from the complementarity of its local and global branches. The local branch improves fine vessel-detail preservation, while the global branch enhances anatomical-context modeling for distinguishing vessels from surrounding interference. The additional gains from E-QKV further indicate that multi-scale QKV generation strengthens global feature representation and improves attention modeling for coronary DSA images with varying vessel sizes, contrast, and noise. Overall, MS-GLWA provides effective global-local context modeling for coronary vessel segmentation.

\subsubsection{Effect of the self-guided feature attention (SGFA) module}
The contribution of SGFA is reflected by comparing Idx 4 and Idx 5 in Table~\ref{table5}. On the right subset, adding SGFA increases Recall from 86.58 to 87.02 and Dice from 87.35 to 87.70, while reducing ASD from 0.8432 to 0.7670 and ACD from 0.8240 to 0.7547. On the left subset, Recall rises from 83.60 to 83.93, Dice from 82.58 to 82.88, ASD decreases from 1.066 to 1.032, and ACD from 1.05 to 1.02.
The improvements obtained with SGFA indicate that global-local context modeling alone is insufficient for optimal coronary DSA segmentation. Although MS-GLWA enhances contextual reasoning, weak distal branches and low-contrast vessels still require dedicated refinement. SGFA addresses this limitation by generating vessel-aware guidance from decoder features and integrating multi-scale contextual information. The reductions in ASD and ACD indicate improved boundary localization and geometric consistency, while the Recall and Dice gains confirm better preservation of weak vessel structures.

\subsubsection{Effect of the proposed boundary-weighted adaptive compound loss (BWACL)}
The contribution of BWACL is shown by comparing Idx 5 and Idx 6 in Table~\ref{table5}. On the right subset, adding BWACL improves Recall from 87.02 to 88.58 and Dice from 87.70 to 88.01, while reducing ASD from 0.7670 to 0.6944 and ACD from 0.7547 to 0.6882. On the left subset, BWACL increases Recall from 83.93 to 85.32, Dice from 82.88 to 82.89, and further reduces ASD from 1.032 to 1.026 and ACD from 1.020 to 1.015. In both subsets, the full configuration (Idx 6) achieves the best overall performance.
These results show that architectural improvements alone are not sufficient; an effective optimization objective is also required to fully exploit the framework. The clear Recall gains on both subsets indicate that BWACL improves vessel recovery, especially for weak and thin vessels, while the reductions in ASD and ACD show that it also enhances boundary precision and structural consistency. This is consistent with the design of BWACL, which explicitly emphasizes thin-vessel boundaries and adaptively balances vessel detection and boundary refinement.

Overall, the results in Table~\ref{table5} confirm that each proposed innovation contributes to the final performance of HTC-SGA Former. From the baseline (Idx 1) to the full model (Idx 6), Recall improves from 85.33 to 88.58 and Dice from 87.14 to 88.01 on the right subset, while ASD decreases from 0.8280 to 0.6944 and ACD from 0.8098 to 0.6882. On the left subset, Recall improves from 81.55 to 85.32, Dice from 82.45 to 82.89, ASD decreases from 1.1126 to 1.0256, and ACD from 1.0845 to 1.0154. These results confirm that MS-GLWA provides effective global-local context modeling, SGFA performs vessel-focused refinement, and BWACL further strengthens optimization toward thin-vessel recovery and boundary precision. Their combination yields the strongest performance on both coronary subsets, demonstrating that the superiority of HTC-SGA Former arises from the complementary interaction between architectural design and loss-function optimization.

\section{Conclusion}
\label{sec:conclusion}
In this study, we proposed HTC-SGA Former, a lightweight hybrid Transformer-CNN framework for coronary DSA vessel segmentation. The framework addresses key challenges associated with weak thin-vessel representation, limited global contextual modeling, severe vessel-background class imbalance, and inaccurate vessel-boundary delineation. Specifically, a CNN encoder preserves fine local vessel morphology, MS-GLWA performs efficient global-local contextual modeling to maintain vessel continuity, SGFA enhances weak vessel responses through vessel-focused refinement, and BWACL improves optimization by emphasizing thin-vessel boundaries and adaptively balancing vessel recovery and boundary refinement.
Experimental results on both right and left coronary artery subsets demonstrate that HTC-SGA Former consistently outperforms 14 state-of-the-art CNN-, attention-, and Transformer-based segmentation methods while maintaining a compact architecture with only 0.81M parameters. In addition, BWACL consistently improves segmentation performance across multiple encoder-decoder backbones, demonstrating its general applicability. These findings indicate that the complementary interaction between architectural design and adaptive optimization enables accurate and parameter-efficient coronary vessel segmentation with improved thin-vessel recovery, vessel continuity preservation, and boundary localization. The proposed framework provides a practical solution for quantitative coronary image analysis and has the potential to support computer-aided diagnosis and future computer-assisted cardiovascular interventions.
Future work will focus on validating HTC-SGA Former on larger multi-center coronary DSA datasets, investigating real-time implementations, and extending the framework to other vascular imaging tasks, including retinal, cerebral, and peripheral vessel segmentation, as well as other angiographic imaging modalities.
\vspace{-10pt}
\\
\\
\\
\textbf{CRediT authorship contribution statement}
\\
\textbf{Rayan Merghani Ahmed}: Conceptualization, Methodology, Software , Validation, Visualization, Writing – original draft. 
\textbf{Marwa Omer Mohammed Omer}: Writing – review \& editing. \textbf{Mohamed Elmanna}: Writing – review \& editing. \textbf{Shijie Li}: Project Administration. \textbf{Bin Li}: Writing – review \& editing. 
 \textbf{Shoujun Zhou}: Conceptualization, Funding acquisition, Project administration, Resources, Supervision, Writing – review \& editing. 
\\ \\
\textbf{Declaration of competing interest}
\\
The authors declare that they have no conflicts of interest.
\\
\\
\textbf{Acknowledgements}
\\
This work was supported by the Shenzhen Medical Research Fund [No. D2404001]; and in part by the Key Research and Development Program of Guangdong Province [No. 2025B1111020001]; the Shenzhen Municipal STIB Key programs [No. CJGJZD20230724093303007, and KJZD2024090310125900]; National Key Laboratory of the CAS on Medical Imaging Science and Technology System, and the Xisike Clinical Oncology Research Foundation[Y-2024AZ(NSCLC)MS-0156].
\\
\\
\textbf{Data availability}
\\
The authors do not have permission to share data.

% \bibliographystyle{unsrtnat}
% \bibliography{cas-refs_u}
\bibliographystyle{elsarticle-harv}
\bibliography{cas-refs}

\begin{thebibliography}{61}
\expandafter\ifx\csname natexlab\endcsname\relax\def\natexlab#1{#1}\fi
\providecommand{\url}[1]{\texttt{#1}}
\providecommand{\href}[2]{#2}
\providecommand{\path}[1]{#1}
\providecommand{\DOIprefix}{doi:}
\providecommand{\ArXivprefix}{arXiv:}
\providecommand{\URLprefix}{URL: }
\providecommand{\Pubmedprefix}{pmid:}
\providecommand{\doi}[1]{\href{http://dx.doi.org/#1}{\path{#1}}}
\providecommand{\Pubmed}[1]{\href{pmid:#1}{\path{#1}}}
\providecommand{\bibinfo}[2]{#2}
\ifx\xfnm\relax \def\xfnm[#1]{\unskip,\space#1}\fi
%Type = Article
\bibitem[{Ahmed et~al.(2026)Ahmed, Iltaf, Elmanna, Zhao, Li, Du, Li and Zhou}]{ahmed2026msa}
\bibinfo{author}{Ahmed, R.M.}, \bibinfo{author}{Iltaf, A.}, \bibinfo{author}{Elmanna, M.}, \bibinfo{author}{Zhao, G.}, \bibinfo{author}{Li, H.}, \bibinfo{author}{Du, Y.}, \bibinfo{author}{Li, B.}, \bibinfo{author}{Zhou, S.}, \bibinfo{year}{2026}.
\newblock \bibinfo{title}{Msa-unet3+: Multi-scale attention unet3+ with new supervised prototypical contrastive loss for coronary dsa image segmentation}.
\newblock \bibinfo{journal}{Biomedical Signal Processing and Control} \bibinfo{volume}{123}, \bibinfo{pages}{110539}.
\newblock \DOIprefix\doi{10.1016/j.bspc.2026.110539}.
%Type = Article
\bibitem[{Algarni et~al.(2022)Algarni, Al-Rezqi, Saeed, Alsaeedi and Ghabban}]{algarni2022multi}
\bibinfo{author}{Algarni, M.}, \bibinfo{author}{Al-Rezqi, A.}, \bibinfo{author}{Saeed, F.}, \bibinfo{author}{Alsaeedi, A.}, \bibinfo{author}{Ghabban, F.}, \bibinfo{year}{2022}.
\newblock \bibinfo{title}{Multi-constraints based deep learning model for automated segmentation and diagnosis of coronary artery disease in x-ray angiographic images}.
\newblock \bibinfo{journal}{PeerJ Computer Science} \bibinfo{volume}{8}, \bibinfo{pages}{e993}.
\newblock \DOIprefix\doi{10.7717/peerj-cs.993}.
%Type = Inproceedings
\bibitem[{Cao et~al.(2022)Cao, Wang, Chen, Jiang, Zhang, Tian and Wang}]{cao2022swin}
\bibinfo{author}{Cao, H.}, \bibinfo{author}{Wang, Y.}, \bibinfo{author}{Chen, J.}, \bibinfo{author}{Jiang, D.}, \bibinfo{author}{Zhang, X.}, \bibinfo{author}{Tian, Q.}, \bibinfo{author}{Wang, M.}, \bibinfo{year}{2022}.
\newblock \bibinfo{title}{Swin-unet: Unet-like pure transformer for medical image segmentation}, in: \bibinfo{booktitle}{European conference on computer vision}, \bibinfo{organization}{Springer}. pp. \bibinfo{pages}{205--218}.
\newblock \DOIprefix\doi{10.1007/978-3-031-25066-8_9}.
%Type = Inproceedings
\bibitem[{Challier et~al.(2025)Challier, Sun, Mahendiran, Senouf, De~Bruyne, Auberson, M{\"u}ller, Fournier, Frossard, Abb{\'e} et~al.}]{challier2025cm}
\bibinfo{author}{Challier, C.}, \bibinfo{author}{Sun, X.}, \bibinfo{author}{Mahendiran, T.}, \bibinfo{author}{Senouf, O.}, \bibinfo{author}{De~Bruyne, B.}, \bibinfo{author}{Auberson, D.}, \bibinfo{author}{M{\"u}ller, O.}, \bibinfo{author}{Fournier, S.}, \bibinfo{author}{Frossard, P.}, \bibinfo{author}{Abb{\'e}, E.}, et~al., \bibinfo{year}{2025}.
\newblock \bibinfo{title}{Cm-unet: A self-supervised learning-based model for coronary artery segmentation in x-ray angiography}, in: \bibinfo{booktitle}{2025 47th Annual International Conference of the IEEE Engineering in Medicine and Biology Society (EMBC)}, \bibinfo{organization}{IEEE}. pp. \bibinfo{pages}{1--7}.
\newblock \DOIprefix\doi{10.1109/embc58623.2025.11253755}.
%Type = Article
\bibitem[{Chang et~al.(2024)Chang, Lin, Wang, Hsu, Wu, Liu and Fann}]{chang2024optimizing}
\bibinfo{author}{Chang, S.S.}, \bibinfo{author}{Lin, C.T.}, \bibinfo{author}{Wang, W.C.}, \bibinfo{author}{Hsu, K.C.}, \bibinfo{author}{Wu, Y.L.}, \bibinfo{author}{Liu, C.H.}, \bibinfo{author}{Fann, Y.C.}, \bibinfo{year}{2024}.
\newblock \bibinfo{title}{Optimizing ensemble u-net architectures for robust coronary vessel segmentation in angiographic images}.
\newblock \bibinfo{journal}{Scientific Reports} \bibinfo{volume}{14}, \bibinfo{pages}{6640}.
\newblock \DOIprefix\doi{10.1038/s41598-024-57198-5}.
%Type = Article
\bibitem[{Chen et~al.(2021)Chen, Lu, Yu, Luo, Adeli, Wang, Lu, Yuille and Zhou}]{chen2021transunet}
\bibinfo{author}{Chen, J.}, \bibinfo{author}{Lu, Y.}, \bibinfo{author}{Yu, Q.}, \bibinfo{author}{Luo, X.}, \bibinfo{author}{Adeli, E.}, \bibinfo{author}{Wang, Y.}, \bibinfo{author}{Lu, L.}, \bibinfo{author}{Yuille, A.L.}, \bibinfo{author}{Zhou, Y.}, \bibinfo{year}{2021}.
\newblock \bibinfo{title}{Transunet: Transformers make strong encoders for medical image segmentation}.
\newblock \bibinfo{journal}{arXiv preprint arXiv:2102.04306} \DOIprefix\doi{10.48550/arXiv.2102.04306}.
%Type = Article
\bibitem[{Deng et~al.(2025)Deng, Li, Liu, Cheng, Fang and Min}]{deng2025multi}
\bibinfo{author}{Deng, H.}, \bibinfo{author}{Li, Y.}, \bibinfo{author}{Liu, X.}, \bibinfo{author}{Cheng, K.}, \bibinfo{author}{Fang, T.}, \bibinfo{author}{Min, X.}, \bibinfo{year}{2025}.
\newblock \bibinfo{title}{Multi-scale dual attention embedded u-shaped network for accurate segmentation of coronary vessels in digital subtraction angiography}.
\newblock \bibinfo{journal}{Medical Physics} \bibinfo{volume}{52}, \bibinfo{pages}{3135--3150}.
\newblock \DOIprefix\doi{10.1002/mp.17618}.
%Type = Article
\bibitem[{Deng et~al.(2024)Deng, Liu, Fang, Li and Min}]{deng2024dfa}
\bibinfo{author}{Deng, H.}, \bibinfo{author}{Liu, X.}, \bibinfo{author}{Fang, T.}, \bibinfo{author}{Li, Y.}, \bibinfo{author}{Min, X.}, \bibinfo{year}{2024}.
\newblock \bibinfo{title}{Dfa-net: Dual multi-scale feature aggregation network for vessel segmentation in x-ray digital subtraction angiography}.
\newblock \bibinfo{journal}{Journal of Big Data} \bibinfo{volume}{11}, \bibinfo{pages}{57}.
\newblock \DOIprefix\doi{10.1186/s40537-024-00904-x}.
%Type = Article
\bibitem[{Fr{\k{a}}k et~al.(2022)Fr{\k{a}}k, Wojtasi{\'n}ska, Lisi{\'n}ska, M{\l}ynarska, Franczyk and Rysz}]{frkak2022pathophysiology}
\bibinfo{author}{Fr{\k{a}}k, W.}, \bibinfo{author}{Wojtasi{\'n}ska, A.}, \bibinfo{author}{Lisi{\'n}ska, W.}, \bibinfo{author}{M{\l}ynarska, E.}, \bibinfo{author}{Franczyk, B.}, \bibinfo{author}{Rysz, J.}, \bibinfo{year}{2022}.
\newblock \bibinfo{title}{Pathophysiology of cardiovascular diseases: new insights into molecular mechanisms of atherosclerosis, arterial hypertension, and coronary artery disease}.
\newblock \bibinfo{journal}{Biomedicines} \bibinfo{volume}{10}, \bibinfo{pages}{1938}.
\newblock \DOIprefix\doi{10.3390/biomedicines10081938}.
%Type = Article
\bibitem[{Fu et~al.(2024a)Fu, Peng, He, Tian, Sun and Wang}]{fu2024hmsu}
\bibinfo{author}{Fu, B.}, \bibinfo{author}{Peng, Y.}, \bibinfo{author}{He, J.}, \bibinfo{author}{Tian, C.}, \bibinfo{author}{Sun, X.}, \bibinfo{author}{Wang, R.}, \bibinfo{year}{2024}a.
\newblock \bibinfo{title}{Hmsu-net: A hybrid multi-scale u-net based on a cnn and transformer for medical image segmentation}.
\newblock \bibinfo{journal}{Computers in Biology and Medicine} \bibinfo{volume}{170}, \bibinfo{pages}{108013}.
\newblock \DOIprefix\doi{10.1016/j.compbiomed.2024.108013}.
%Type = Article
\bibitem[{Fu et~al.(2024b)Fu, Fu, Lu, Yan, Fei and Han}]{fu2024robust}
\bibinfo{author}{Fu, Z.}, \bibinfo{author}{Fu, Z.}, \bibinfo{author}{Lu, C.}, \bibinfo{author}{Yan, J.}, \bibinfo{author}{Fei, J.}, \bibinfo{author}{Han, H.}, \bibinfo{year}{2024}b.
\newblock \bibinfo{title}{Robust implementation of foreground extraction and vessel segmentation for x-ray coronary angiography image sequence}.
\newblock \bibinfo{journal}{Pattern Recognition} \bibinfo{volume}{145}, \bibinfo{pages}{109926}.
\newblock \DOIprefix\doi{10.1016/j.patcog.2023.109926}.
%Type = Article
\bibitem[{Gao et~al.(2023)Gao, Li, Yang and Liu}]{gao2023multi}
\bibinfo{author}{Gao, G.}, \bibinfo{author}{Li, J.}, \bibinfo{author}{Yang, L.}, \bibinfo{author}{Liu, Y.}, \bibinfo{year}{2023}.
\newblock \bibinfo{title}{\protect{A multi-scale global attention network for blood vessel segmentation from fundus images}}.
\newblock \bibinfo{journal}{Measurement} \bibinfo{volume}{222}, \bibinfo{pages}{113553}.
\newblock \DOIprefix\doi{10.1016/j.measurement.2023.113553}.
%Type = Article
\bibitem[{Gao et~al.(2026)Gao, Wang, Ai, Shang, Song, Fan, Xiao, Wang and Yang}]{gao2026iterative}
\bibinfo{author}{Gao, Y.}, \bibinfo{author}{Wang, Y.}, \bibinfo{author}{Ai, D.}, \bibinfo{author}{Shang, F.}, \bibinfo{author}{Song, H.}, \bibinfo{author}{Fan, J.}, \bibinfo{author}{Xiao, D.}, \bibinfo{author}{Wang, Y.}, \bibinfo{author}{Yang, J.}, \bibinfo{year}{2026}.
\newblock \bibinfo{title}{Iterative joint learning integrating temporal and geometric information for vessel segmentation in x-ray coronary angiography}.
\newblock \bibinfo{journal}{Medical Physics} \bibinfo{volume}{53}, \bibinfo{pages}{e70317}.
\newblock \DOIprefix\doi{10.1002/mp.70317}.
%Type = Article
\bibitem[{He et~al.(2023)He, Wang, Li, Du, Xia and Fu}]{he2023h2former}
\bibinfo{author}{He, A.}, \bibinfo{author}{Wang, K.}, \bibinfo{author}{Li, T.}, \bibinfo{author}{Du, C.}, \bibinfo{author}{Xia, S.}, \bibinfo{author}{Fu, H.}, \bibinfo{year}{2023}.
\newblock \bibinfo{title}{H2former: An efficient hierarchical hybrid transformer for medical image segmentation}.
\newblock \bibinfo{journal}{IEEE Transactions on Medical Imaging} \bibinfo{volume}{42}, \bibinfo{pages}{2763--2775}.
\newblock \DOIprefix\doi{10.1109/tmi.2023.3264513}.
%Type = Inproceedings
\bibitem[{He et~al.(2016)He, Zhang, Ren and Sun}]{he2016deep}
\bibinfo{author}{He, K.}, \bibinfo{author}{Zhang, X.}, \bibinfo{author}{Ren, S.}, \bibinfo{author}{Sun, J.}, \bibinfo{year}{2016}.
\newblock \bibinfo{title}{Deep residual learning for image recognition}, in: \bibinfo{booktitle}{Proceedings of the IEEE conference on computer vision and pattern recognition}, pp. \bibinfo{pages}{770--778}.
\newblock \DOIprefix\doi{10.1109/cvpr.2016.90}.
%Type = Article
\bibitem[{Huang et~al.(2025)Huang, Luo, Wei, He, Shao, Zeng and Zhang}]{huang2025deep}
\bibinfo{author}{Huang, B.}, \bibinfo{author}{Luo, Y.}, \bibinfo{author}{Wei, G.}, \bibinfo{author}{He, S.}, \bibinfo{author}{Shao, Y.}, \bibinfo{author}{Zeng, X.}, \bibinfo{author}{Zhang, Q.}, \bibinfo{year}{2025}.
\newblock \bibinfo{title}{Deep learning model for coronary artery segmentation and quantitative stenosis detection in angiographic images}.
\newblock \bibinfo{journal}{Medical Physics} \bibinfo{volume}{52}, \bibinfo{pages}{e17970}.
\newblock \DOIprefix\doi{10.1002/mp.17970}.
%Type = Article
\bibitem[{Huang et~al.(2022)Huang, Deng, Li, Yuan and Fu}]{huang2022missformer}
\bibinfo{author}{Huang, X.}, \bibinfo{author}{Deng, Z.}, \bibinfo{author}{Li, D.}, \bibinfo{author}{Yuan, X.}, \bibinfo{author}{Fu, Y.}, \bibinfo{year}{2022}.
\newblock \bibinfo{title}{Missformer: An effective transformer for 2d medical image segmentation}.
\newblock \bibinfo{journal}{IEEE transactions on medical imaging} \bibinfo{volume}{42}, \bibinfo{pages}{1484--1494}.
\newblock \DOIprefix\doi{10.1109/TMI.2022.3230943}.
%Type = Article
\bibitem[{Huang et~al.(2024)Huang, Gong and Zhang}]{huang2024hst}
\bibinfo{author}{Huang, X.}, \bibinfo{author}{Gong, H.}, \bibinfo{author}{Zhang, J.}, \bibinfo{year}{2024}.
\newblock \bibinfo{title}{Hst-mrf: heterogeneous swin transformer with multi-receptive field for medical image segmentation}.
\newblock \bibinfo{journal}{IEEE Journal of Biomedical and Health Informatics} \bibinfo{volume}{28}, \bibinfo{pages}{4048--4061}.
\newblock \DOIprefix\doi{10.1109/jbhi.2024.3397047}.
%Type = Inproceedings
\bibitem[{Li et~al.(2020)Li, Tan, Chen, Luo, Gao, Jia and Wang}]{li2020attention}
\bibinfo{author}{Li, C.}, \bibinfo{author}{Tan, Y.}, \bibinfo{author}{Chen, W.}, \bibinfo{author}{Luo, X.}, \bibinfo{author}{Gao, Y.}, \bibinfo{author}{Jia, X.}, \bibinfo{author}{Wang, Z.}, \bibinfo{year}{2020}.
\newblock \bibinfo{title}{Attention unet++: A nested attention-aware u-net for liver ct image segmentation}, in: \bibinfo{booktitle}{2020 IEEE international conference on image processing (ICIP)}, \bibinfo{organization}{IEEE}. pp. \bibinfo{pages}{345--349}.
\newblock \DOIprefix\doi{10.1109/icip40778.2020.9190761}.
%Type = Article
\bibitem[{Li et~al.(2023a)Li, Ye, Zhang, Wu, Berhane, Deng and Shi}]{li2023cpftransformer}
\bibinfo{author}{Li, J.}, \bibinfo{author}{Ye, J.}, \bibinfo{author}{Zhang, R.}, \bibinfo{author}{Wu, Y.}, \bibinfo{author}{Berhane, G.S.}, \bibinfo{author}{Deng, H.}, \bibinfo{author}{Shi, H.}, \bibinfo{year}{2023}a.
\newblock \bibinfo{title}{Cpftransformer: transformer fusion context pyramid medical image segmentation network}.
\newblock \bibinfo{journal}{Frontiers in Neuroscience} \bibinfo{volume}{17}, \bibinfo{pages}{1288366}.
\newblock \DOIprefix\doi{10.3389/fnins.2023.1288366}.
%Type = Article
\bibitem[{Li et~al.(2023b)Li, Fang, Yang, Su, Zhu and Yu}]{li2023transu2}
\bibinfo{author}{Li, X.}, \bibinfo{author}{Fang, X.}, \bibinfo{author}{Yang, G.}, \bibinfo{author}{Su, S.}, \bibinfo{author}{Zhu, L.}, \bibinfo{author}{Yu, Z.}, \bibinfo{year}{2023}b.
\newblock \bibinfo{title}{Transu$^2$-net: An effective medical image segmentation framework based on transformer and u$^2$-net}.
\newblock \bibinfo{journal}{IEEE Journal of Translational Engineering in Health and Medicine} \bibinfo{volume}{11}, \bibinfo{pages}{441--450}.
\newblock \DOIprefix\doi{10.1109/jtehm.2023.3289990}.
%Type = Inproceedings
\bibitem[{Lin et~al.(2017)Lin, Goyal, Girshick, He and Doll{\'a}r}]{lin2017focal}
\bibinfo{author}{Lin, T.Y.}, \bibinfo{author}{Goyal, P.}, \bibinfo{author}{Girshick, R.}, \bibinfo{author}{He, K.}, \bibinfo{author}{Doll{\'a}r, P.}, \bibinfo{year}{2017}.
\newblock \bibinfo{title}{Focal loss for dense object detection}, in: \bibinfo{booktitle}{Proceedings of the IEEE international conference on computer vision}, pp. \bibinfo{pages}{2980--2988}.
\newblock \DOIprefix\doi{10.1109/iccv.2017.324}.
%Type = Article
\bibitem[{Liu et~al.(2024)Liu, Wang, Wang, Hu, Wang and Ge}]{liu2024imff}
\bibinfo{author}{Liu, M.}, \bibinfo{author}{Wang, Y.}, \bibinfo{author}{Wang, L.}, \bibinfo{author}{Hu, S.}, \bibinfo{author}{Wang, X.}, \bibinfo{author}{Ge, Q.}, \bibinfo{year}{2024}.
\newblock \bibinfo{title}{\protect{IMFF-Net: An integrated multi-scale feature fusion network for accurate retinal vessel segmentation from fundus images}}.
\newblock \bibinfo{journal}{Biomedical Signal Processing and Control} \bibinfo{volume}{91}, \bibinfo{pages}{105980}.
\newblock \DOIprefix\doi{10.1016/j.bspc.2024.105980}.
%Type = Article
\bibitem[{Liu et~al.(2022)Liu, Yang, Tian, Cao, Pan, Xu, Jin and Gao}]{liu2022full}
\bibinfo{author}{Liu, W.}, \bibinfo{author}{Yang, H.}, \bibinfo{author}{Tian, T.}, \bibinfo{author}{Cao, Z.}, \bibinfo{author}{Pan, X.}, \bibinfo{author}{Xu, W.}, \bibinfo{author}{Jin, Y.}, \bibinfo{author}{Gao, F.}, \bibinfo{year}{2022}.
\newblock \bibinfo{title}{Full-resolution network and dual-threshold iteration for retinal vessel and coronary angiograph segmentation}.
\newblock \bibinfo{journal}{IEEE journal of biomedical and health informatics} \bibinfo{volume}{26}, \bibinfo{pages}{4623--4634}.
\newblock \DOIprefix\doi{10.1109/jbhi.2022.3188710}.
%Type = Article
\bibitem[{Liu et~al.(2023)Liu, Zhu, Xin, Zhang, Yang and Xu}]{liu2023mestrans}
\bibinfo{author}{Liu, Y.}, \bibinfo{author}{Zhu, Y.}, \bibinfo{author}{Xin, Y.}, \bibinfo{author}{Zhang, Y.}, \bibinfo{author}{Yang, D.}, \bibinfo{author}{Xu, T.}, \bibinfo{year}{2023}.
\newblock \bibinfo{title}{Mestrans: Multi-scale embedding spatial transformer for medical image segmentation}.
\newblock \bibinfo{journal}{Computer Methods and Programs in Biomedicine} \bibinfo{volume}{233}, \bibinfo{pages}{107493}.
\newblock \DOIprefix\doi{10.1016/j.cmpb.2023.107493}.
%Type = Inproceedings
\bibitem[{Liu et~al.(2021)Liu, Lin, Cao, Hu, Wei, Zhang, Lin and Guo}]{liu2021swin}
\bibinfo{author}{Liu, Z.}, \bibinfo{author}{Lin, Y.}, \bibinfo{author}{Cao, Y.}, \bibinfo{author}{Hu, H.}, \bibinfo{author}{Wei, Y.}, \bibinfo{author}{Zhang, Z.}, \bibinfo{author}{Lin, S.}, \bibinfo{author}{Guo, B.}, \bibinfo{year}{2021}.
\newblock \bibinfo{title}{Swin transformer: Hierarchical vision transformer using shifted windows}, in: \bibinfo{booktitle}{Proceedings of the IEEE/CVF international conference on computer vision}, pp. \bibinfo{pages}{10012--10022}.
\newblock \DOIprefix\doi{10.1109/iccv48922.2021.00986}.
%Type = Article
\bibitem[{Mardani et~al.(2025)Mardani, Maghooli and Farokhi}]{mardani2025segmentation}
\bibinfo{author}{Mardani, K.}, \bibinfo{author}{Maghooli, K.}, \bibinfo{author}{Farokhi, F.}, \bibinfo{year}{2025}.
\newblock \bibinfo{title}{Segmentation of coronary arteries from x-ray angiographic images using density based spatial clustering of applications with noise (dbscan)}.
\newblock \bibinfo{journal}{Biomedical Signal Processing and Control} \bibinfo{volume}{101}, \bibinfo{pages}{107175}.
\newblock \DOIprefix\doi{10.1016/j.bspc.2024.107175}.
%Type = Inproceedings
\bibitem[{Milletari et~al.(2016)Milletari, Navab and Ahmadi}]{milletari2016v}
\bibinfo{author}{Milletari, F.}, \bibinfo{author}{Navab, N.}, \bibinfo{author}{Ahmadi, S.A.}, \bibinfo{year}{2016}.
\newblock \bibinfo{title}{V-net: Fully convolutional neural networks for volumetric medical image segmentation}, in: \bibinfo{booktitle}{2016 fourth international conference on 3D vision (3DV)}, \bibinfo{organization}{Ieee}. pp. \bibinfo{pages}{565--571}.
\newblock \DOIprefix\doi{10.1109/3dv.2016.79}.
%Type = Article
\bibitem[{Moccia et~al.(2018)Moccia, De~Momi, El~Hadji and Mattos}]{moccia2018blood}
\bibinfo{author}{Moccia, S.}, \bibinfo{author}{De~Momi, E.}, \bibinfo{author}{El~Hadji, S.}, \bibinfo{author}{Mattos, L.S.}, \bibinfo{year}{2018}.
\newblock \bibinfo{title}{Blood vessel segmentation algorithms—review of methods, datasets and evaluation metrics}.
\newblock \bibinfo{journal}{Computer methods and programs in biomedicine} \bibinfo{volume}{158}, \bibinfo{pages}{71--91}.
\newblock \DOIprefix\doi{10.1016/j.cmpb.2018.02.001}.
%Type = Article
\bibitem[{Molenaar et~al.(2025)Molenaar, Hebbo, Selder, Shekiladze, Sandesara, Nicholson, Asselbergs, Ahmad, Gold, Sakr et~al.}]{molenaar2025deep}
\bibinfo{author}{Molenaar, M.A.}, \bibinfo{author}{Hebbo, E.}, \bibinfo{author}{Selder, J.L.}, \bibinfo{author}{Shekiladze, N.}, \bibinfo{author}{Sandesara, P.B.}, \bibinfo{author}{Nicholson, W.J.}, \bibinfo{author}{Asselbergs, F.W.}, \bibinfo{author}{Ahmad, S.}, \bibinfo{author}{Gold, D.A.}, \bibinfo{author}{Sakr, S.M.}, et~al., \bibinfo{year}{2025}.
\newblock \bibinfo{title}{Deep learning--based segmentation of coronary arteries and stenosis detection in x-ray coronary angiography}.
\newblock \bibinfo{journal}{JACC: Advances} \bibinfo{volume}{4}, \bibinfo{pages}{102360}.
\newblock \DOIprefix\doi{10.1016/j.jacadv.2025.102360}.
%Type = Article
\bibitem[{Nobre~Menezes et~al.(2023)Nobre~Menezes, Silva, Silva, Rodrigues, Guerreiro, Guedes, Santos, Oliveira and Pinto}]{nobre2023coronary}
\bibinfo{author}{Nobre~Menezes, M.}, \bibinfo{author}{Silva, J.L.}, \bibinfo{author}{Silva, B.}, \bibinfo{author}{Rodrigues, T.}, \bibinfo{author}{Guerreiro, C.}, \bibinfo{author}{Guedes, J.P.}, \bibinfo{author}{Santos, M.O.}, \bibinfo{author}{Oliveira, A.L.}, \bibinfo{author}{Pinto, F.J.}, \bibinfo{year}{2023}.
\newblock \bibinfo{title}{Coronary x-ray angiography segmentation using artificial intelligence: a multicentric validation study of a deep learning model}.
\newblock \bibinfo{journal}{The international journal of cardiovascular imaging} \bibinfo{volume}{39}, \bibinfo{pages}{1385--1396}.
\newblock \DOIprefix\doi{10.1007/s10554-023-02839-5}.
%Type = Article
\bibitem[{Oktay et~al.(2018)Oktay, Schlemper, Folgoc, Lee, Heinrich, Misawa, Mori, McDonagh, Hammerla, Kainz et~al.}]{oktay2018attention}
\bibinfo{author}{Oktay, O.}, \bibinfo{author}{Schlemper, J.}, \bibinfo{author}{Folgoc, L.L.}, \bibinfo{author}{Lee, M.}, \bibinfo{author}{Heinrich, M.}, \bibinfo{author}{Misawa, K.}, \bibinfo{author}{Mori, K.}, \bibinfo{author}{McDonagh, S.}, \bibinfo{author}{Hammerla, N.Y.}, \bibinfo{author}{Kainz, B.}, et~al., \bibinfo{year}{2018}.
\newblock \bibinfo{title}{Attention u-net: Learning where to look for the pancreas}.
\newblock \bibinfo{journal}{arXiv preprint arXiv:1804.03999} \DOIprefix\doi{10.48550/arXiv.1804.03999}.
%Type = Article
\bibitem[{Pagliaro et~al.(2020)Pagliaro, Cannata, Stefanini and Bolognese}]{pagliaro2020myocardial}
\bibinfo{author}{Pagliaro, B.R.}, \bibinfo{author}{Cannata, F.}, \bibinfo{author}{Stefanini, G.G.}, \bibinfo{author}{Bolognese, L.}, \bibinfo{year}{2020}.
\newblock \bibinfo{title}{Myocardial ischemia and coronary disease in heart failure}.
\newblock \bibinfo{journal}{Heart failure reviews} \bibinfo{volume}{25}, \bibinfo{pages}{53--65}.
\newblock \DOIprefix\doi{10.1007/s10741-019-09831-z}.
%Type = Article
\bibitem[{Pan et~al.(2021)Pan, Li, Su, Tay, Tran and Chan}]{pan2021coronary}
\bibinfo{author}{Pan, L.S.}, \bibinfo{author}{Li, C.W.}, \bibinfo{author}{Su, S.F.}, \bibinfo{author}{Tay, S.Y.}, \bibinfo{author}{Tran, Q.V.}, \bibinfo{author}{Chan, W.P.}, \bibinfo{year}{2021}.
\newblock \bibinfo{title}{Coronary artery segmentation under class imbalance using a u-net based architecture on computed tomography angiography images}.
\newblock \bibinfo{journal}{Scientific reports} \bibinfo{volume}{11}, \bibinfo{pages}{14493}.
\newblock \DOIprefix\doi{10.1038/s41598-021-93889-z}.
%Type = Article
\bibitem[{Park et~al.(2023)Park, Kweon, Kim, Back, Chae, Roh, Kang, Lee, Ahn, Kang et~al.}]{park2023selective}
\bibinfo{author}{Park, J.}, \bibinfo{author}{Kweon, J.}, \bibinfo{author}{Kim, Y.I.}, \bibinfo{author}{Back, I.}, \bibinfo{author}{Chae, J.}, \bibinfo{author}{Roh, J.H.}, \bibinfo{author}{Kang, D.Y.}, \bibinfo{author}{Lee, P.H.}, \bibinfo{author}{Ahn, J.M.}, \bibinfo{author}{Kang, S.J.}, et~al., \bibinfo{year}{2023}.
\newblock \bibinfo{title}{Selective ensemble methods for deep learning segmentation of major vessels in invasive coronary angiography}.
\newblock \bibinfo{journal}{Medical physics} \bibinfo{volume}{50}, \bibinfo{pages}{7822--7839}.
\newblock \DOIprefix\doi{10.1002/mp.16554}.
%Type = Article
\bibitem[{Peng et~al.(2024)Peng, Wang, Pedersoli and Desrosiers}]{peng2024boundary}
\bibinfo{author}{Peng, J.}, \bibinfo{author}{Wang, P.}, \bibinfo{author}{Pedersoli, M.}, \bibinfo{author}{Desrosiers, C.}, \bibinfo{year}{2024}.
\newblock \bibinfo{title}{Boundary-aware information maximization for self-supervised medical image segmentation}.
\newblock \bibinfo{journal}{Medical Image Analysis} \bibinfo{volume}{94}, \bibinfo{pages}{103150}.
\newblock \DOIprefix\doi{10.1016/j.media.2024.103150}.
%Type = Article
\bibitem[{Pu et~al.(2023)Pu, Zhang, Qian, Zeng, Li, Zhang, Zhou and Zhao}]{pu2023semi}
\bibinfo{author}{Pu, Y.}, \bibinfo{author}{Zhang, Q.}, \bibinfo{author}{Qian, C.}, \bibinfo{author}{Zeng, Q.}, \bibinfo{author}{Li, N.}, \bibinfo{author}{Zhang, L.}, \bibinfo{author}{Zhou, S.}, \bibinfo{author}{Zhao, G.}, \bibinfo{year}{2023}.
\newblock \bibinfo{title}{Semi-supervised segmentation of coronary dsa using mixed networks and multi-strategies}.
\newblock \bibinfo{journal}{Computers in Biology and Medicine} \bibinfo{volume}{156}, \bibinfo{pages}{106493}.
\newblock \DOIprefix\doi{10.1016/j.compbiomed.2022.106493}.
%Type = Article
\bibitem[{Qiu et~al.(2022)Qiu, Zhang and Song}]{qiu2022dynamic}
\bibinfo{author}{Qiu, M.}, \bibinfo{author}{Zhang, C.}, \bibinfo{author}{Song, Z.}, \bibinfo{year}{2022}.
\newblock \bibinfo{title}{Dynamic boundary-insensitive loss for magnetic resonance medical image segmentation}.
\newblock \bibinfo{journal}{Medical physics} \bibinfo{volume}{49}, \bibinfo{pages}{1739--1753}.
\newblock \DOIprefix\doi{10.1002/mp.15386}.
%Type = Inproceedings
\bibitem[{Ronneberger et~al.(2015)Ronneberger, Fischer and Brox}]{ronneberger2015u}
\bibinfo{author}{Ronneberger, O.}, \bibinfo{author}{Fischer, P.}, \bibinfo{author}{Brox, T.}, \bibinfo{year}{2015}.
\newblock \bibinfo{title}{U-net: Convolutional networks for biomedical image segmentation}, in: \bibinfo{booktitle}{Medical image computing and computer-assisted intervention--MICCAI 2015: 18th international conference, Munich, Germany, October 5-9, 2015, proceedings, part III 18}, \bibinfo{organization}{Springer}. pp. \bibinfo{pages}{234--241}.
\newblock \DOIprefix\doi{10.1007/978-3-319-24574-4_28}.
%Type = Article
\bibitem[{Samuel and Veeramalai(2021)}]{samuel2021vssc}
\bibinfo{author}{Samuel, P.M.}, \bibinfo{author}{Veeramalai, T.}, \bibinfo{year}{2021}.
\newblock \bibinfo{title}{Vssc net: vessel specific skip chain convolutional network for blood vessel segmentation}.
\newblock \bibinfo{journal}{Computer methods and programs in biomedicine} \bibinfo{volume}{198}, \bibinfo{pages}{105769}.
\newblock \DOIprefix\doi{10.1016/j.cmpb.2020.105769}.
%Type = Article
\bibitem[{Shen et~al.(2023)Shen, Chen, Tong, Jiang and Ning}]{shen2023dbcu}
\bibinfo{author}{Shen, Y.}, \bibinfo{author}{Chen, Z.}, \bibinfo{author}{Tong, J.}, \bibinfo{author}{Jiang, N.}, \bibinfo{author}{Ning, Y.}, \bibinfo{year}{2023}.
\newblock \bibinfo{title}{Dbcu-net: deep learning approach for segmentation of coronary angiography images}.
\newblock \bibinfo{journal}{The International Journal of Cardiovascular Imaging} \bibinfo{volume}{39}, \bibinfo{pages}{1571--1579}.
\newblock \DOIprefix\doi{10.1007/s10554-023-02849-3}.
%Type = Inproceedings
\bibitem[{Tang et~al.(2024)Tang, Ding, Quan, Wang, Ning and Zhou}]{tang2024cmunext}
\bibinfo{author}{Tang, F.}, \bibinfo{author}{Ding, J.}, \bibinfo{author}{Quan, Q.}, \bibinfo{author}{Wang, L.}, \bibinfo{author}{Ning, C.}, \bibinfo{author}{Zhou, S.K.}, \bibinfo{year}{2024}.
\newblock \bibinfo{title}{Cmunext: An efficient medical image segmentation network based on large kernel and skip fusion}, in: \bibinfo{booktitle}{2024 IEEE International Symposium on Biomedical Imaging (ISBI)}, \bibinfo{organization}{IEEE}. pp. \bibinfo{pages}{1--5}.
\newblock \DOIprefix\doi{10.1109/isbi56570.2024.10635609}.
%Type = Inproceedings
\bibitem[{Tang et~al.(2023a)Tang, Wang, Ning, Xian and Ding}]{tang2023cmu}
\bibinfo{author}{Tang, F.}, \bibinfo{author}{Wang, L.}, \bibinfo{author}{Ning, C.}, \bibinfo{author}{Xian, M.}, \bibinfo{author}{Ding, J.}, \bibinfo{year}{2023}a.
\newblock \bibinfo{title}{Cmu-net: a strong convmixer-based medical ultrasound image segmentation network}, in: \bibinfo{booktitle}{2023 IEEE 20th international symposium on biomedical imaging (ISBI)}, \bibinfo{organization}{IEEE}. pp. \bibinfo{pages}{1--5}.
\newblock \DOIprefix\doi{10.1109/isbi53787.2023.10230609}.
%Type = Article
\bibitem[{Tang et~al.(2023b)Tang, Duan, Sun, Zeng, Zhang and Yao}]{tang2023combined}
\bibinfo{author}{Tang, Z.}, \bibinfo{author}{Duan, J.}, \bibinfo{author}{Sun, Y.}, \bibinfo{author}{Zeng, Y.}, \bibinfo{author}{Zhang, Y.}, \bibinfo{author}{Yao, X.}, \bibinfo{year}{2023}b.
\newblock \bibinfo{title}{A combined deformable model and medical transformer algorithm for medical image segmentation}.
\newblock \bibinfo{journal}{Medical \& Biological Engineering \& Computing} \bibinfo{volume}{61}, \bibinfo{pages}{129--137}.
\newblock \DOIprefix\doi{10.1007/s11517-022-02702-0}.
%Type = Article
\bibitem[{Wang et~al.(2024)Wang, Zhou, Gao, Qin, Wang, Sun and Yu}]{wang2024coronary}
\bibinfo{author}{Wang, G.}, \bibinfo{author}{Zhou, P.}, \bibinfo{author}{Gao, H.}, \bibinfo{author}{Qin, Z.}, \bibinfo{author}{Wang, S.}, \bibinfo{author}{Sun, J.}, \bibinfo{author}{Yu, H.}, \bibinfo{year}{2024}.
\newblock \bibinfo{title}{Coronary vessel segmentation in coronary angiography with a multi-scale u-shaped transformer incorporating boundary aggregation and topology preservation}.
\newblock \bibinfo{journal}{Physics in Medicine \& Biology} \bibinfo{volume}{69}, \bibinfo{pages}{025012}.
\newblock \DOIprefix\doi{10.1088/1361-6560/ad0b63}.
%Type = Article
\bibitem[{Wang et~al.(2025)Wang, Qi, Liu, Guo, Lv and Liang}]{wang2025dpgnet}
\bibinfo{author}{Wang, H.}, \bibinfo{author}{Qi, Y.}, \bibinfo{author}{Liu, W.}, \bibinfo{author}{Guo, K.}, \bibinfo{author}{Lv, W.}, \bibinfo{author}{Liang, Z.}, \bibinfo{year}{2025}.
\newblock \bibinfo{title}{Dpgnet: A boundary-aware medical image segmentation framework via uncertainty perception}.
\newblock \bibinfo{journal}{IEEE Journal of Biomedical and Health Informatics} \DOIprefix\doi{10.1109/jbhi.2025.3601025}.
%Type = Article
\bibitem[{Wang et~al.(2022)Wang, Chen, Ji, Fan and Li}]{wang2022boundary}
\bibinfo{author}{Wang, R.}, \bibinfo{author}{Chen, S.}, \bibinfo{author}{Ji, C.}, \bibinfo{author}{Fan, J.}, \bibinfo{author}{Li, Y.}, \bibinfo{year}{2022}.
\newblock \bibinfo{title}{Boundary-aware context neural network for medical image segmentation}.
\newblock \bibinfo{journal}{Medical image analysis} \bibinfo{volume}{78}, \bibinfo{pages}{102395}.
\newblock \DOIprefix\doi{10.1016/j.media.2022.102395}.
%Type = Inproceedings
\bibitem[{Woo et~al.(2018)Woo, Park, Lee and Kweon}]{woo2018cbam}
\bibinfo{author}{Woo, S.}, \bibinfo{author}{Park, J.}, \bibinfo{author}{Lee, J.Y.}, \bibinfo{author}{Kweon, I.S.}, \bibinfo{year}{2018}.
\newblock \bibinfo{title}{Cbam: Convolutional block attention module}, in: \bibinfo{booktitle}{Proceedings of the European conference on computer vision (ECCV)}, pp. \bibinfo{pages}{3--19}.
\newblock \DOIprefix\doi{10.1007/978-3-030-01234-2_1}.
%Type = Article
\bibitem[{Wu et~al.(2024)Wu, Min, Gai, Huang, Geng, Wang and Chen}]{wu2024hd}
\bibinfo{author}{Wu, H.}, \bibinfo{author}{Min, W.}, \bibinfo{author}{Gai, D.}, \bibinfo{author}{Huang, Z.}, \bibinfo{author}{Geng, Y.}, \bibinfo{author}{Wang, Q.}, \bibinfo{author}{Chen, R.}, \bibinfo{year}{2024}.
\newblock \bibinfo{title}{Hd-former: A hierarchical dependency transformer for medical image segmentation}.
\newblock \bibinfo{journal}{Computers in Biology and Medicine} \bibinfo{volume}{178}, \bibinfo{pages}{108671}.
\newblock \DOIprefix\doi{10.1016/j.compbiomed.2024.108671}.
%Type = Article
\bibitem[{Wu et~al.(2025)Wu, Dai, Chen, Huang, Xiao, Ma and Ouyang}]{wu2025adaptive}
\bibinfo{author}{Wu, W.}, \bibinfo{author}{Dai, T.}, \bibinfo{author}{Chen, Z.}, \bibinfo{author}{Huang, X.}, \bibinfo{author}{Xiao, J.}, \bibinfo{author}{Ma, F.}, \bibinfo{author}{Ouyang, R.}, \bibinfo{year}{2025}.
\newblock \bibinfo{title}{Adaptive patch contrast for weakly supervised semantic segmentation}.
\newblock \bibinfo{journal}{Engineering Applications of Artificial Intelligence} \bibinfo{volume}{139}, \bibinfo{pages}{109626}.
\newblock \DOIprefix\doi{10.1016/j.engappai.2024.109626}.
%Type = Article
\bibitem[{Xia et~al.(2020)Xia, Rook, Pelgrim, Sidorenkov, Wisselink, Van~Bolhuis, van Ooijen, Guo, Oudkerk, Groen et~al.}]{xia2020early}
\bibinfo{author}{Xia, C.}, \bibinfo{author}{Rook, M.}, \bibinfo{author}{Pelgrim, G.J.}, \bibinfo{author}{Sidorenkov, G.}, \bibinfo{author}{Wisselink, H.J.}, \bibinfo{author}{Van~Bolhuis, J.N.}, \bibinfo{author}{van Ooijen, P.M.}, \bibinfo{author}{Guo, J.}, \bibinfo{author}{Oudkerk, M.}, \bibinfo{author}{Groen, H.}, et~al., \bibinfo{year}{2020}.
\newblock \bibinfo{title}{Early imaging biomarkers of lung cancer, copd and coronary artery disease in the general population: rationale and design of the imalife (imaging in lifelines) study: C. xia et al.}
\newblock \bibinfo{journal}{European journal of epidemiology} \bibinfo{volume}{35}, \bibinfo{pages}{75--86}.
\newblock \DOIprefix\doi{10.1007/s10654-019-00519-0}.
%Type = Article
\bibitem[{Xie et~al.(2023)Xie, Zhang, Pan, Xie, Shao, Zhao and An}]{xie2023canet}
\bibinfo{author}{Xie, X.}, \bibinfo{author}{Zhang, W.}, \bibinfo{author}{Pan, X.}, \bibinfo{author}{Xie, L.}, \bibinfo{author}{Shao, F.}, \bibinfo{author}{Zhao, W.}, \bibinfo{author}{An, J.}, \bibinfo{year}{2023}.
\newblock \bibinfo{title}{Canet: Context aware network with dual-stream pyramid for medical image segmentation}.
\newblock \bibinfo{journal}{Biomedical Signal Processing and Control} \bibinfo{volume}{81}, \bibinfo{pages}{104437}.
\newblock \DOIprefix\doi{10.1016/j.bspc.2022.104437}.
%Type = Article
\bibitem[{Yang et~al.(2019)Yang, Kweon, Roh, Lee, Kang, Park, Kim, Yang, Hur, Kang et~al.}]{yang2019deep}
\bibinfo{author}{Yang, S.}, \bibinfo{author}{Kweon, J.}, \bibinfo{author}{Roh, J.H.}, \bibinfo{author}{Lee, J.H.}, \bibinfo{author}{Kang, H.}, \bibinfo{author}{Park, L.J.}, \bibinfo{author}{Kim, D.J.}, \bibinfo{author}{Yang, H.}, \bibinfo{author}{Hur, J.}, \bibinfo{author}{Kang, D.Y.}, et~al., \bibinfo{year}{2019}.
\newblock \bibinfo{title}{Deep learning segmentation of major vessels in x-ray coronary angiography}.
\newblock \bibinfo{journal}{Scientific reports} \bibinfo{volume}{9}, \bibinfo{pages}{16897}.
\newblock \DOIprefix\doi{10.1038/s41598-019-53254-7}.
%Type = Article
\bibitem[{Zeng et~al.(2019)Zeng, Liu, Xiao, Li, Jiang, Feng and Guo}]{zeng2019automatic}
\bibinfo{author}{Zeng, Y.}, \bibinfo{author}{Liu, X.}, \bibinfo{author}{Xiao, N.}, \bibinfo{author}{Li, Y.}, \bibinfo{author}{Jiang, Y.}, \bibinfo{author}{Feng, J.}, \bibinfo{author}{Guo, S.}, \bibinfo{year}{2019}.
\newblock \bibinfo{title}{Automatic diagnosis based on spatial information fusion feature for intracranial aneurysm}.
\newblock \bibinfo{journal}{IEEE transactions on medical imaging} \bibinfo{volume}{39}, \bibinfo{pages}{1448--1458}.
\newblock \DOIprefix\doi{10.1109/tmi.2019.2951439}.
%Type = Article
\bibitem[{Zhang et~al.(2022)Zhang, Gao, Zhang, Hau and Zhang}]{zhang2022progressive}
\bibinfo{author}{Zhang, H.}, \bibinfo{author}{Gao, Z.}, \bibinfo{author}{Zhang, D.}, \bibinfo{author}{Hau, W.K.}, \bibinfo{author}{Zhang, H.}, \bibinfo{year}{2022}.
\newblock \bibinfo{title}{Progressive perception learning for main coronary segmentation in x-ray angiography}.
\newblock \bibinfo{journal}{IEEE Transactions on Medical Imaging} \bibinfo{volume}{42}, \bibinfo{pages}{864--879}.
\newblock \DOIprefix\doi{10.1109/tmi.2022.3219126}.
%Type = Article
\bibitem[{Zhang et~al.(2023a)Zhang, Zhong, Li, Liu, Liu, Ji, Li and Wu}]{zhang2023bcu}
\bibinfo{author}{Zhang, H.}, \bibinfo{author}{Zhong, X.}, \bibinfo{author}{Li, G.}, \bibinfo{author}{Liu, W.}, \bibinfo{author}{Liu, J.}, \bibinfo{author}{Ji, D.}, \bibinfo{author}{Li, X.}, \bibinfo{author}{Wu, J.}, \bibinfo{year}{2023}a.
\newblock \bibinfo{title}{Bcu-net: Bridging convnext and u-net for medical image segmentation}.
\newblock \bibinfo{journal}{Computers in Biology and Medicine} \bibinfo{volume}{159}, \bibinfo{pages}{106960}.
\newblock \DOIprefix\doi{10.1016/j.compbiomed.2023.106960}.
%Type = Article
\bibitem[{Zhang et~al.(2024)Zhang, Wang, Wang, Saif and Wassan}]{zhang2024cidn}
\bibinfo{author}{Zhang, M.}, \bibinfo{author}{Wang, H.}, \bibinfo{author}{Wang, L.}, \bibinfo{author}{Saif, A.}, \bibinfo{author}{Wassan, S.}, \bibinfo{year}{2024}.
\newblock \bibinfo{title}{Cidn: A context interactive deep network with edge-aware for x-ray angiography images segmentation}.
\newblock \bibinfo{journal}{Alexandria Engineering Journal} \bibinfo{volume}{87}, \bibinfo{pages}{201--212}.
\newblock \DOIprefix\doi{10.1016/j.aej.2023.12.034}.
%Type = Article
\bibitem[{Zhang et~al.(2023b)Zhang, Gao, Zhou, He, Xia, Peng, Lou, Zhou, Tang and Chen}]{zhang2023centerline}
\bibinfo{author}{Zhang, Y.}, \bibinfo{author}{Gao, Y.}, \bibinfo{author}{Zhou, G.}, \bibinfo{author}{He, J.}, \bibinfo{author}{Xia, J.}, \bibinfo{author}{Peng, G.}, \bibinfo{author}{Lou, X.}, \bibinfo{author}{Zhou, S.}, \bibinfo{author}{Tang, H.}, \bibinfo{author}{Chen, Y.}, \bibinfo{year}{2023}b.
\newblock \bibinfo{title}{Centerline-supervision multi-task learning network for coronary angiography segmentation}.
\newblock \bibinfo{journal}{Biomedical Signal Processing and Control} \bibinfo{volume}{82}, \bibinfo{pages}{104510}.
\newblock \DOIprefix\doi{10.1016/j.bspc.2022.104510}.
%Type = Article
\bibitem[{Zhong et~al.(2025)Zhong, Tian, Xie, Liu, Ou, Tian and Zhang}]{zhong2025pmfsnet}
\bibinfo{author}{Zhong, J.}, \bibinfo{author}{Tian, W.}, \bibinfo{author}{Xie, Y.}, \bibinfo{author}{Liu, Z.}, \bibinfo{author}{Ou, J.}, \bibinfo{author}{Tian, T.}, \bibinfo{author}{Zhang, L.}, \bibinfo{year}{2025}.
\newblock \bibinfo{title}{\protect{PMFSNet: Polarized multi-scale feature self-attention network for lightweight medical image segmentation}}.
\newblock \bibinfo{journal}{Computer Methods and Programs in Biomedicine} , \bibinfo{pages}{108611}.
%Type = Article
\bibitem[{Zhu et~al.(2021)Zhu, Cheng, Wang, Chen and Lu}]{zhu2021coronary}
\bibinfo{author}{Zhu, X.}, \bibinfo{author}{Cheng, Z.}, \bibinfo{author}{Wang, S.}, \bibinfo{author}{Chen, X.}, \bibinfo{author}{Lu, G.}, \bibinfo{year}{2021}.
\newblock \bibinfo{title}{Coronary angiography image segmentation based on pspnet}.
\newblock \bibinfo{journal}{Computer Methods and Programs in Biomedicine} \bibinfo{volume}{200}, \bibinfo{pages}{105897}.
\newblock \DOIprefix\doi{10.1016/j.cmpb.2020.105897}.
%Type = Article
\bibitem[{Zongwei et~al.(2022)Zongwei, Rahman, Nima and Jianming}]{Zongwei_Rahman_Nima_Jianming_2022}
\bibinfo{author}{Zongwei, Z.}, \bibinfo{author}{Rahman, Siddiquee, M.M.}, \bibinfo{author}{Nima, T.}, \bibinfo{author}{Jianming, L.}, \bibinfo{year}{2022}.
\newblock \bibinfo{title}{Unet++: Redesigning skip connections to exploit multiscale features in image segmentation}.
\newblock \bibinfo{journal}{arXiv (Cornell University)} \URLprefix \url{https://arxiv.org/abs/1912.05074}.

\end{thebibliography}

\end{document}